\pdfoutput=1
\documentclass[11pt]{article}

\usepackage[]{EMNLP2023}

\usepackage{times}
\usepackage{latexsym}

\usepackage[T1]{fontenc}
\usepackage[utf8]{inputenc}

\usepackage{microtype}

\usepackage{inconsolata}

\usepackage{graphicx}
\usepackage{subfigure}
\usepackage{tabularx,booktabs}
\usepackage{enumitem}
\usepackage{amsmath}
\usepackage{amssymb}
\usepackage{multirow}
\usepackage{algorithm,algorithmic}
\usepackage{pifont}
\usepackage{colortbl}
\newcommand{\ouralg}{AIMMerging} 

\title{AIMMerging: Adaptive Iterative Model Merging Using Training Trajectories for Language Model Continual Learning}

\author{Yujie Feng$^{1,2}$\thanks{~ Equal contribution.}\enspace, Jian Li$^{1}$\footnotemark[1]\enspace, Xiaoyu Dong$^{2}$, Pengfei Xu$^{1}$, Xiaohui Zhou$^{1}$, \textbf{Yujia Zhang}$^{1}$ \\ \textbf{Zexin Lu}$^{2}$\textbf{,} \textbf{Yasha Wang}$^{3}$, \textbf{Alan Zhao}$^{1}$\thanks{ ~ Corresponding author.}\enspace, \textbf{Xu Chu}$^{3}$\footnotemark[2]\enspace, \textbf{Xiao-Ming Wu}$^{2}$\footnotemark[2] \\
$^1$Al Technology Center of OVB, Tencent, China \\
$^2$The Hong Kong Polytechnic University, Hong Kong S.A.R.
$^3$Peking University, China \\
yuujiefeng@tencent.com, xiao-ming.wu@polyu.edu.hk 
}

\begin{document}
\maketitle
\begin{abstract}
Continual learning (CL) is essential for deploying large language models (LLMs) in dynamic real-world environments without the need for costly retraining. Recent model merging-based methods have attracted significant attention, but they still struggle to effectively manage the trade-off between learning new knowledge and preventing forgetting, a challenge largely stemming from suboptimal number of merges and merging frequency. In this paper, we introduce Adaptive Iterative Model Merging ({\ouralg}), a novel CL framework that utilizes learning and forgetting signals from the training trajectory to dynamically monitor the model's training status. Guided by dynamic monitoring, the training trajectory-guided merge controller adaptively determines the timing and frequency of iterative fusion, while the rehearsal-based knowledge fusion module computes the merging weights and executes the fusion. Comprehensive experiments on three CL benchmarks with various model sizes (from 770M to 13B) demonstrate that {\ouralg} achieves significant performance improvements over existing state-of-the-art methods, with an average relative improvement of 80\% and 59\% on FWT and BWT, respectively. The source code\footnote{\url{https://github.com/WoodScene/AimMerging}} is provided for reproducibility.

\end{abstract}

\section{Introduction}

% Endowing the continual learning (CL) ability for large language models (LLMs) to learn different tasks sequentially is essential for their deployment in the ever-changing environments without extensive retraining \cite{wang2024comprehensive, jiang2024interpretable, yu2024recent, chang2024survey, chen2024entity}. 
% However, two major challenges arise in CL: (1) Catastrophic Forgetting (CF)~\cite{mccloskey1989catastrophic}, which refers to the loss of previously learned knowledge when acquiring new tasks, and (2) Knowledge Transfer (KT)~\cite{ke2021achieving}, which involves leveraging knowledge from new, related tasks to enhance performance on prior tasks, and vice versa.
% Striking a balance between retaining previous knowledge and excelling in new tasks is vital for success.

% Modifications from Jian Li
% ---
Continual learning (CL) is vital for the effective deployment of large language models (LLMs) in evolving environments, allowing them to sequentially acquire new knowledge and circumventing the necessity of costly retraining \cite{liao2025data, eskandar2025star, wang2024malora, jiang2024interpretable, yu2024recent, chang2024survey}. However, the core challenge in CL lies in effectively balancing the retention of previously learned knowledge (mitigating catastrophic forgetting, CF \cite{mccloskey1989catastrophic}) with the acquisition of new knowledge (facilitating knowledge transfer, KT \cite{ke2021achieving}). Successfully managing this inherent trade-off is vital for practical deployment.
% ---

\begin{figure}[t]
  \centering
  \includegraphics[width=1.0\linewidth]{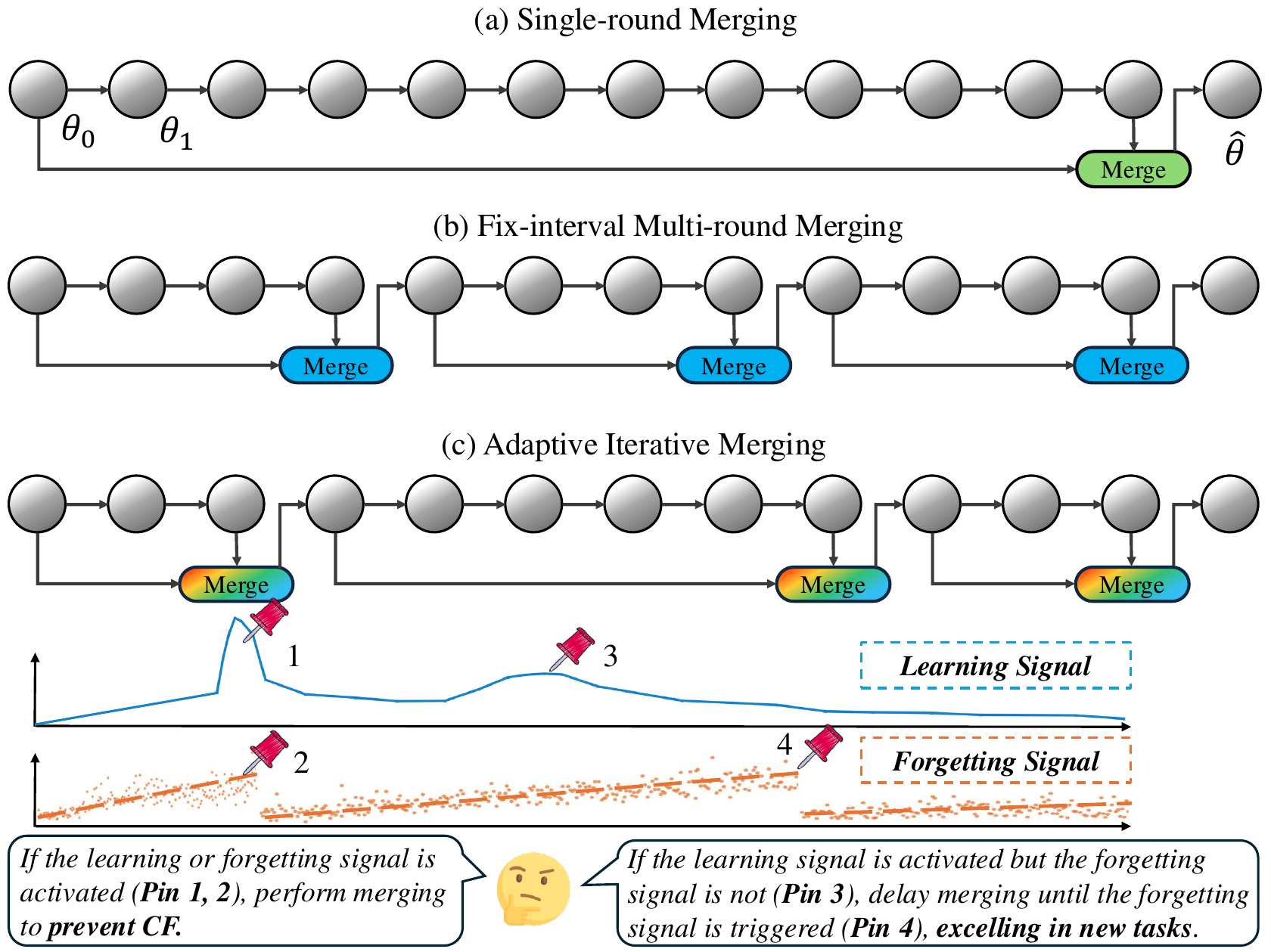}
  \caption{
Illustration of three different model merging strategies.
%single-round, fix-interval multi-round, and the proposed adaptive iterative merging. 
Guided by the learning and forgetting signals extracted from the training trajectory, our {\ouralg} adaptively adjusts the merging intervals and frequency, thereby enhancing CL performance.
  }
  \label{fig:intro}
\end{figure}

% Recently, model merging-based methods have emerged as effective solutions for CL in LLMs \cite{dou-etal-2024-loramoe, wan2024knowledge, yadav2024survey}, due to their strong ability to facilitate knowledge transfer.
% Traditional approaches typically involve a single model merging operation between the pre- and post-training models, with two main strategies: the global merging and the fine-grained merging, as shown in Fig. 1(a).
% In contrast to single-round merging methods, \citet{feng2025recurrent} proposed a recurrent knowledge identification and fusion framework that performs merging after every fixed number of training iterations, demonstrating that multiple rounds of merging can further enhance performance by leveraging valuable intermediate states during training (Fig. 1b).

% Modifications from Jian Li
% ---

Recent model merging methods \cite{dou-etal-2024-loramoe, wan2024knowledge, yadav2024survey} have gained prominence for CL, largely due to their capacity for KT. Traditional approaches typically involve a single-round merge, commonly applied between pre- and post-training models, using global or fine-grained strategies (Figure \ref{fig:intro}(a)). Departing from single-round methods, \citet{feng2025recurrent} proposed a recurrent framework that merges models iteratively after fixed training steps (Figure \ref{fig:intro}(b)), showing that leveraging intermediate training states through multiple merges can enhance performance.
% --- 

%This approach reveals the huge potential of multi-round merging frameworks and provides further space for exploration and optimization in future research. Inspired by the strong performance of multi-round merging techniques, a natural question arises:

% Modifications from Jian Li
% ---
This multi-round merging paradigm reveals significant potential and highlights the importance of optimizing the merging process. Inspired by these promising results, a critical question emerges:
% ---

%\textbf{\textit{How can we determine the optimal moments for merging during model training, and how do we choose the appropriate number of merges to further improve performance?}}

% Modifications from Jian Li
% ---
\textbf{\textit{How can we determine the optimal timing and frequency of merging during training to further enhance performance?}}
% ---

%To address this, we introduce \textbf{A}daptive \textbf{I}terative \textbf{M}odel \textbf{Merging} (\textbf{\ouralg}) for continual learning, which leverages the training trajectories through a \textit{\textbf{learning signal}} for new knowledge and a \textit{\textbf{forgetting signal}} for historical knowledge to enable dynamic and real-time monitoring of the model's training status.

% Modifications from Jian Li
% ---
To this end, we propose a novel CL framework called \textbf{A}daptive \textbf{I}terative \textbf{M}odel \textbf{Merging} (\textbf{\ouralg}). It achieves dynamic monitoring of the training status by innovatively employing \textit{\textbf{learning}} and \textit{\textbf{forgetting signals}} extracted from the training trajectory.
% ---
%Based on these signals, {\ouralg} consists of two key modules: the \textbf{Training Trajectory-guided Merge Controller}, which governs the timing, frequency, and intervals of model merging, and the \textbf{Rehearsal-based Knowledge Fusion Module}, which performs the actual merging operation using a global merging strategy to achieve a better balance between retaining previous knowledge and excelling in new tasks.
% Modifications from Jian Li
% ---
Based on these signals, {\ouralg} consists of two key modules: the \textbf{Training Trajectory-guided Merge Controller}, responsible for adaptively scheduling the timing and frequency of model merging, and the \textbf{Rehearsal-based Knowledge Fusion Module}, which performs the global merging operation. 
More specifically, the \textit{\textbf{learning signal}} is quantified by the change in model parameters across training steps, reflecting the model's acquisition progress for new knowledge. Analysis of its trend via a sliding window identifies periods of rapid learning (peak in Figure \ref{fig:newknowledge}) or slow convergence (downward trend in Figure \ref{fig:newknowledge}). The \textit{\textbf{forgetting signal}}, on the other hand, is derived from the loss on historical data, offering real-time insight into the extent of CF. It is triggered when the historical loss exceeds a predefined threshold or shows a notable rise, signifying potential knowledge loss.
These signals guide the merge controller with distinct functions. The learning signal, typically measured after a merge, helps determine the next merging interval, thereby influencing the overall merging frequency. In contrast, the forgetting signal is continuously monitored during training. It serves as a critical, real-time trigger, prompting an immediate merge when significant forgetting of historical knowledge occurs.

%These signals guide the merge controller with distinct functions. The learning signal, derived from the training trajectory between merges, is analyzed to assess the learning progress of new knowledge. This analysis, typically performed after a merge, informs the determination of the subsequent merging interval, thus influencing the overall merging frequency. Conversely, the \textit{\textbf{forgetting signal}} is continuously monitored during training iterations. It serves as a critical, real-time trigger, indicating the immediate necessity for a merge when the model exhibits significant forgetting of historical knowledge.
% ---

% Based on the different states of these two signals, our Training Trajectory-guided Merge Controller dynamically and adaptively adjusts the model's merging timing and frequency to improve performance. In general, when the learning signal is activated, the Merge Controller shortens the interval to the next merge, preventing excessive forgetting of previous knowledge due to the rapid learning of new knowledge. If, by the expected merging time, the forgetting signal has not been activated, the merge interval is extended until the forgetting signal is triggered, allowing the model to focus more on learning new knowledge. However, if the forgetting signal is repeatedly activated before reaching the expected merging point, a merge will be triggered earlier to avoid significant forgetting of historical knowledge. For a more detailed implementation, please refer to Section 3.
% Modifications from Jian Li
% ---
Leveraging the dynamic monitoring from the learning and forgetting signals, the \textbf{training trajectory-guided merge controller} adaptively determines the merging schedule by interpreting their interplay.
Based on the findings from our preliminary study (see Section \ref{Preliminary}), the controller increases the merging frequency to proactively mitigate CF when the learning signal indicates a rapid learning phase or the forgetting signal is activated.
Conversely, when the learning signal indicates a slow convergence phase or the forgetting signal remains inactive, the controller reduces the merging frequency, allowing the model to focus more on learning new knowledge. Through this interaction, our method strikes a better balance between retaining previous knowledge and excelling in new tasks.
The \textbf{rehearsal-based knowledge fusion module} executes the global merge operation, utilizing the relative importance weights derived from the learning and forgetting signals to merge new and historical knowledge effectively.
Extensive experiments demonstrate the strong performance of our method in addressing CL challenges.

Our main contributions are summarized as:
% Modifications from Jian Li
% ---
% Our main contributions are summarized as follows:
% ---

\begin{itemize}[leftmargin=*,itemsep=2pt,topsep=0pt,parsep=0pt]
\item 
%To the best of our knowledge, {\ouralg} is the first to analyze the model's status during fine-tuning, specifically the learning of new knowledge and forgetting of historical knowledge, based on training trajectories for CL. This allows for the adaptive adjustment of the model's merging strategy to enable iterative knowledge fusion.

% To the best of our knowledge, {\ouralg} is the first to leverage the training trajectory to analyze the model's learning and forgetting states during fine-tuning in CL. Guided by a learning signal for new knowledge and a forgetting signal for historical knowledge, {\ouralg} enables adaptive adjustments to the timing, frequency, and weights of the iterative merging strategy.
% Modifications from Jian Li
% ---
We propose a novel adaptive iterative model merging \textbf{framework} ({\ouralg}) for CL. To the best of our knowledge, {\ouralg} is the first to leverage the training trajectory by extracting learning and forgetting signals to dynamically monitor the model's state and guiding adaptive scheduling of iterative model merging.
%guiding the adaptive scheduling and execution of iterative model merging.
% ---

\item 
% We develop novel training trajectory-guided merger controller and rehearsal-based knowledge fusion techniques.
% Modifications from Jian Li
% ---
%We develop two new training trajectory-guided merge controller and rehearsal-based knowledge fusion \textbf{techniques}.
We introduce two novel \textbf{techniques}: the training trajectory-guided merge controller and the rehearsal-based knowledge fusion module.
%two key novel components within {\ouralg}: the \textbf{Training Trajectory-guided Merge Controller}, which adaptively determines the merging timing and frequency, and the \textbf{Rehearsal-based Knowledge Fusion Module}, which calculates signal-informed merging weights using rehearsal data and performs the fusion. 
% ---

\item 
% Extensive evaluation validates the effectiveness of {\ouralg} in addressing CL challenges.
% Modifications from Jian Li
% ---
%Extensive \textbf{evaluation} on three CL benchmarks using four different backbones (from 770M to 13B) demonstrates that {\ouralg} consistently outperforms state-of-the-art methods, effectively mitigating CF and promoting KT.
% ---

%Extensive \textbf{evaluation} on three CL benchmarks using four different backbones (from 770M to 13B) demonstrates {\ouralg}'s strong knowledge transfer capability, with 80\% (from -2.5\% to -0.5\%) and 59\% (from -4.9\% to -2.0\%) relative improvement on the FWT and BWT, respectively, outperforming previous state-of-the-art methods.
Extensive \textbf{evaluation} on three CL benchmarks utilizing four backbones (from 770M to 13B) demonstrates that {\ouralg} significantly enhances knowledge transfer capabilities, achieving an average relative improvement of 80\% (from -2.5\% to -0.5\%) in FWT  and 59\% (from -4.9\% to -2.0\%) in BWT , surpassing previous state-of-the-art methods.

\end{itemize}

\section{Preliminary Study}
\label{Preliminary}

%In this section, we conduct experimental explorations to guide the development of an appropriate and optimal adaptive merging strategy. These explorations include: (i) investigating the dynamic changes in new knowledge learning and historical knowledge forgetting within the model’s training states, and (ii) examining the impact of model merging during training on both new knowledge learning and the forgetting of historical knowledge. These analyses provide the motivation for improving our adaptive merging strategy and offer valuable insights for optimizing multi-round merging-based methods.
In this section, we conduct two key analysis:
(i) investigating the dynamic changes in the model's training states regarding new knowledge acquisition and historical knowledge forgetting, and (ii) examining the impact of model merging during training on both new and historical knowledge. These analyses provide valuable insights for optimizing the adaptive merging strategy.

We first define the problem and introduce the relevant concepts for better clarity. All experiments in this section are conducted on long sequence benchmarks using T5-large.

\paragraph{Problem Formulation}
%\subsection{Continual Learning Setup}
Continual learning aims to progressively accumulate knowledge from a sequence of tasks $\{\mathcal{T}_1, \ldots, \mathcal{T}_K\}$. Each task $\mathcal{T}_k$ includes a distinct dataset $\mathcal{D}_k = \left\{ \left( x_i^k, y_i^k \right) \right\}_{i=1}^{N_k}$ of size $N_k$, where $x_i^k \in \mathcal{X}_k$ and $y_i^k \in \mathcal{Y}_k$.
The model, parameterized by $\Theta$, is trained sequentially on these tasks to minimize the following objective:
% \begin{equation}
% \max_{\Theta} \sum_{k=1}^{K} \sum_{x,y \in \mathcal{D}_k} \log p_{\Theta}(y \mid x)
% \end{equation}
\begin{equation}
\mathcal{L} = \mathbb{E}_{(x, y) \sim \bigcup_{k=1}^K \mathcal{D}_k} \left[ -\log p_\Theta(y \mid x) \right]
\end{equation}

In this work, we consider a practical scenario where a small portion of data from previous tasks is stored in a memory buffer to facilitate the CL process. 
Specifically, we randomly store $\left| \mathcal{M} \right|$ samples from each task $\mathcal{T}_i$ in memory $\mathcal{M}_i$. During training, the model is jointly optimized on the new task data $\mathcal{D}_k$ and the memory buffer $\mathcal{M}_{<k}$.

\paragraph{Notation}
We consider a pre-trained model $\theta \in \mathbb{R}^n$ with $n$ parameters.
After training on task $\mathcal{T}_{k-1}$, the model are denoted as $\theta^{k-1}$.
Fine-tuning on a new task $\mathcal{T}_k$ produces updated parameters $\theta^k$.
The difference $\tau^k = \theta^k - \theta^{k-1}$, referred to as the \textit{task vector} or \textit{training residual} \cite{ilharco2023editing}, represents task-specific parameter updates.

For traditional single-round merging methods, a merging function \( f_{\text{merge}} \) is typically used to combine the model \( \theta^{k-1} \) and the fine-tuned model \( \theta^k \) to obtain the final model: $\hat{\theta}^k = f_{\text{merge}}(\theta^{k-1}, \theta^k)$.
In contrast, multi-round model merging methods perform merges during the training process. Specifically, assuming the total number of training iterations is \( J \), \( \theta^{k-1}_j \) represents the model's parameters at the \( j \)-th iteration. 
For example, if the interval between two consecutive merges is \( S \) (e.g., 100 training iterations), the merged model is represented as: $\hat{\theta}^{k-1}_{j+S} = f^{\prime}_{\text{merge}}(\theta^{k-1}_j, \theta^{k-1}_{j+S})$.
The model is then further trained based on $\hat{\theta}^{k-1}_{j+S}$.

%\subsection{Analysis of New Knowledge Acquisition and Historical Knowledge Forgetting During Model Training}
\subsection{Analysis of Knowledge Acquisition and Forgetting}
\begin{figure}[t]
  \centering
  \includegraphics[width=1\linewidth]{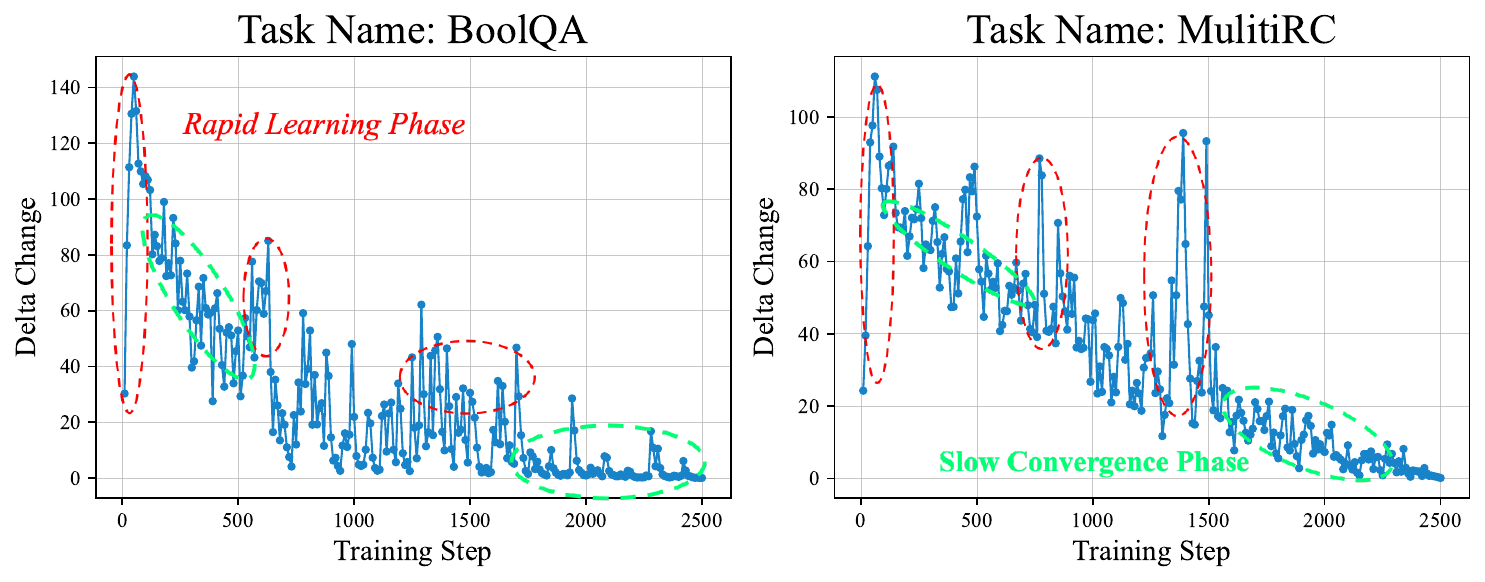}
  \caption{Parameter change for new knowledge acquisition during training.
  }
  \label{fig:newknowledge}
\end{figure}

\paragraph{New Knowledge Acquisition} 
We measure the model's learning state for new knowledge by summing the absolute values of parameter changes within a fixed training interval, such as every 10 steps (Fig. \ref{fig:newknowledge}). In the early stages of training, the parameter changes are large and show an upward trend, indicating the rapid learning phase. As training progresses, these changes decrease, signaling a slow convergence phase. Interestingly, peaks may reappear, suggesting the model revisits unlearned or challenging knowledge.

%We assess the model's learning state for new knowledge by summing the absolute values of all parameter changes within a fixed training interval, such as every 10 training steps, as shown in Fig. \ref{fig:newknowledge}. In the early stages of training, parameter changes are larger and exhibit an upward trend, which we define as the rapid learning phase for new knowledge. As training progresses, these changes diminish, entering a slow convergence phase. This aligns with the intuitive notion that models are typically unstable at the beginning and stabilize over time. Interestingly, after some training, one or more peaks may emerge, indicating the model has re-entered the rapid learning phase, likely revisiting previously unlearned or challenging knowledge.

Based on two learning scenarios, we conducted two comparative experiments: (1) increasing merging frequency during the rapid learning phase, and (2) increasing merging frequency during the slow convergence phase. As shown in Table \ref{tbl:pre_learn}, the results reveal that increasing merging frequency during the rapid learning phase improves performance by preventing excessive accumulation of new knowledge. However, merging during the convergence phase leads to a decline in performance, likely due to redundancy in the stable model. This insight is valuable for refining the learning signal strategy in our method.
%Based on the two distinct new knowledge learning scenarios, we conducted two comparative experiments: (1) increasing the merging frequency during the model's rapid learning phase, and (2) increasing the merging frequency during the model's slow convergence phase. As shown in Table \ref{tbl:pre_learn}, the findings reveal that increasing the merging frequency during the rapid learning phase improves performance, as more frequent merging helps prevent excessive accumulation of new knowledge at each step. Conversely, increasing the merging frequency during the convergence phase leads to a performance decline, likely due to the model reaching a stable state, where further merging introduces unnecessary redundancy. This insight is crucial for refining the learning signal strategy in our method.

\begin{table}[t]
\centering
\scalebox{0.85}{
\begin{tabular}{lccc}
\toprule
Merging Strategy & OP &  FWT & BWT\\
\midrule
\rowcolor[gray]{1}
\rule{0pt}{6pt} Fix Interval  & 78.3  & -3.4  &  -2.7 \\
\midrule

\rule{0pt}{8pt} Slow Convergence Phase$^+$ & 77.9  & -3.8 & -3.0  \\
\rule{0pt}{8pt} Rapid Learning Phase$^+$  & \textbf{78.5} & \textbf{-2.7}  & \textbf{-1.9}  \\

\bottomrule
\end{tabular}}
\caption{The impact of different merging strategies on performance. ``$^+$'' indicates increased merging frequency during the corresponding phases.
}
\label{tbl:pre_learn}
\end{table}

\begin{figure}[t]
  \centering
  \includegraphics[width=1\linewidth]{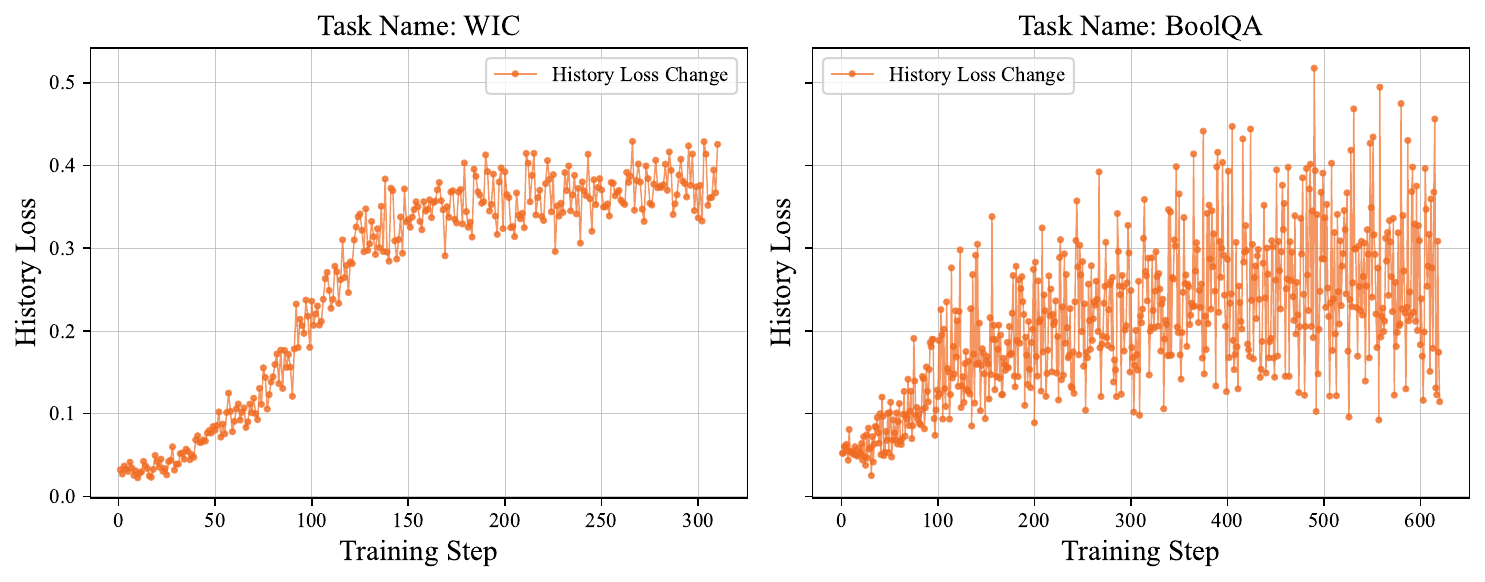}
  \caption{Changes in historical loss during training.
  }
  \label{fig:historicalknowledge}
\end{figure}

\paragraph{Forgetting of Historical Knowledge}
During training, we sample a batch of memory data from the buffer, feeding it into the model for loss calculation without gradient updates. This allows us to monitor historical knowledge loss, as shown in Fig. \ref{fig:historicalknowledge}. The loss for historical knowledge increases as new knowledge is learned, aligning with expectations. When the loss exceeds a predefined threshold, the forgetting signal is triggered, prompting merging to mitigate forgetting.
%During training with the latest task, we extract a batch of memory data from the memory buffer and feed it into the model. This batch is used exclusively for loss calculation, with no gradient updates, allowing us to monitor the loss change associated with historical knowledge, as shown in Fig. \ref{fig:historicalknowledge}. From the figure, we observe that the loss for historical knowledge increases as new knowledge is learned, which aligns with the intuitive expectation. Based on this observation, we treat the change in historical task loss as a forgetting signal. When the loss exceeds a predefined threshold, the forgetting signal is triggered, prompting the model to perform merging in order to mitigate forgetting.

\subsection{Impact of Model Merging on New and Historical Knowledge}
\begin{figure}[t]
  \centering
  \includegraphics[width=1\linewidth]{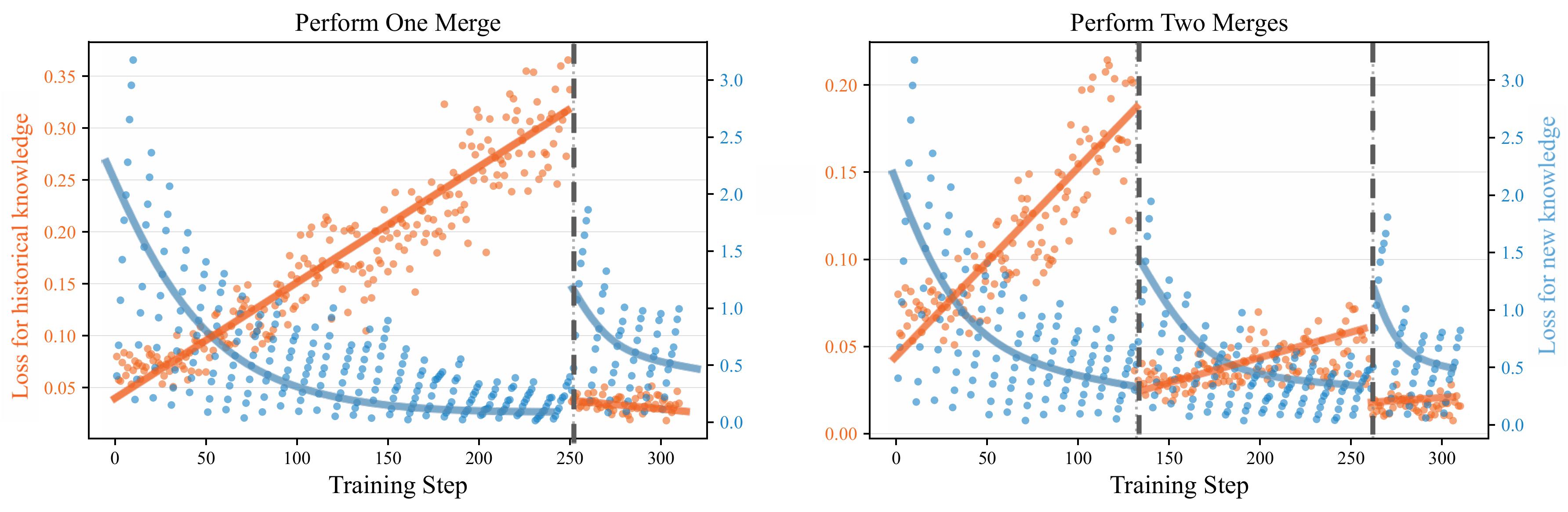}
  \caption{Loss change for new and historical knowledge after model merges during training.
  }
  \label{fig:mergetime}
\end{figure}

We perform one or two merges during training to observe the changes in loss for both new and historical knowledge (Figure \ref{fig:mergetime}). The results show that after each merge, historical knowledge loss decreases significantly, demonstrating effective mitigation of CF. However, the loss for new tasks increases, indicating that merging may interfere with learning new knowledge. Thus, selecting the appropriate merging timing is key to balancing new knowledge acquisition and historical knowledge retention.
%We perform one or two merges during model training to observe the changes in loss for both new and historical knowledge, as shown in Fig. \ref{fig:mergetime}. The results reveal that after each merge, the loss for historical tasks decreases significantly, indicating that model merging effectively mitigates CF. However, the loss for the new task increases following each merge, suggesting that merging may interfere with the learning of new knowledge. Therefore, selecting the appropriate merging timing is crucial for balancing the impact of new knowledge acquisition with the preservation of historical knowledge.

\section{Proposed Method: {\ouralg}}
\label{Method}

\paragraph{Overview}

{\ouralg} employs two signals, the \textit{\textbf{learning signal}} and the \textit{\textbf{forgetting signal}}, to monitor the model’s training state.
%specifically tracking the learning of new knowledge and the degree of forgetting of historical knowledge in real-time.
Based on the real-time fluctuations of these two signals, our approach reconfigures the training process into multiple iterative model merging cycles, guided by two core components:
(i) \textit{\textbf{Training Trajectory-guided Merge Controller}}: adaptively selects the timing of model merges and dynamically adjusts the intervals between subsequent merges.
(ii) \textit{\textbf{Rehearsal-based Knowledge Fusion}}: responsible for generating merging weights and applying memory-replay techniques to integrate new and historical knowledge.

\subsection{The Design of Merge Controller}
%Assume the current task is \(\mathcal{T}_k\), and the model's initial parameter state is denoted by \(\theta_{k-1}\). After training on data $D_k$ for $j$ iterations, the parameter state is represented as \(\theta_j^{k-1}\)\footnote{For simplicity, we omit the superscript \(k-1\) in the following descriptions.}.
Assume the current task is \(\mathcal{T}_k\), and \(\theta_j^{k-1}\) represents the model's state after $j$ training iterations \footnote{For simplicity, we omit the superscript \(k-1\) in the following descriptions.}.

\paragraph{Merge Controller with Learning Signal}  
The core function of the learning signal is to dynamically adjust the merging interval $S$ based on the model's current state for new knowledge.
Assume the $b$-th merge is scheduled at the $j$-th training iteration, and the interval since the $(b{-}1)$-th merge is $S_b$.
The task vector capturing the parameter update between these two successive merges is defined as:
\begin{equation}
\tau_{b} = \theta_j - \theta_{j - S_b}
\end{equation}

By summing the absolute values of the task vector elements, we obtain a measure of the model's learning state for the new task, expressed as:
\begin{equation}
\Lambda_{b} = \sum_{i=1}^{n} |\tau_{b}^i| \, / \, S_b
\end{equation}
where \(\tau_{b}^i\) is the \(i\)-th element of the task vector. Dividing by \(S_b\) normalizes the value, allowing for fair comparison across intervals of varying lengths.

We use \(\Lambda\) to assess the model's learning state for new knowledge. By comparing the current value \(\Lambda_{b}\) with the previous one \(\Lambda_{b-1}\), we observe the parameter change trend and adjust the merging interval from \(S_b\) to \(S_{b+1}\).

However, considering only the trend between two consecutive values may cause the learning signal to be overly sensitive to short-term fluctuations. To address this, we adopt a sliding window approach to analyze parameter change trends across multiple historical points and capture a more reliable overall trajectory.
Specifically, we maintain a list to record the historical values of $\Lambda$, denoted as $\mathcal{H} = [\Lambda_1, \Lambda_2, \dots, \Lambda_{b-1}]$. Given a sliding window length $L_w$, we compare the trends between consecutive entries, i.e., between $\Lambda_{b}$ and $\Lambda_{b-1}$, $\Lambda_{b-1}$ and $\Lambda_{b-2}$, $\dots$, up to $\Lambda_{b-L_w+1}$ and $\Lambda_{b-L_w}$.

If upward trends dominate, indicating rapid learning phrase for new knowledge, we reduce the merging interval based on the magnitude of parameter changes (Case 1). If downward trends dominate, indicating slow convergence phase, we increase the merging interval (Case 3). 
If they are balanced, we keep the current interval (Case 2).
The adjustment strategy is defined as:
\begin{equation}
S_{b+1} =
\begin{cases}
\max(S_{\min}, S_b / \gamma_{\text{learn}}^{-}), & \text{(Case 1)} \\
S_b, & \text{(Case 2)} \\
\min(S_{\max}, S_b \cdot \gamma_{\text{learn}}^{+}), & \text{(Case 3)}
\end{cases}
\end{equation}
where $S_{\min}$ and $S_{\max}$ denote the minimum and maximum allowed merging intervals, and $\gamma_{\text{learn}}$ is a step-size adjustment factor.
A cold-start phase is introduced at the beginning of training, during which no adjustments are made, lasting the length of the sliding window with an initial interval $S_{init}$.
%At the start of training, a cold-start phase is introduced, during which no merging interval adjustment occurs. This phase typically lasts the length of the sliding window, with an initial merging interval $S_{init}$.

\paragraph{Merge Controller with Both Learning and Forgetting Signals}
%Without the forgetting signal, the model dynamically adjusts the next merging interval \(S\) solely on the learning signal, performing merges at the corresponding training iterations and adaptively refining the interval for the next merge. However, this approach only considers the model's learning state for new knowledge and does not analyze the forgetting of historical knowledge, resulting in an incomplete analysis.
Relying only on the learning signal, the model adjusts the merging interval $S$ based on the learning state for new knowledge, but this neglects the forgetting of historical knowledge, leading to a suboptimal merging strategy.

To address this, we further integrate the forgetting signal $\mathcal{F}$ to assist the controller adjust the merging strategy by considering both new and historical knowledge.
The strategy triggers earlier or delayed merges depending on the forgetting signal, optimizing the merging interval.
%A more comprehensive optimization strategy is as follows: depending on the state of the forgetting signal, the controller may either trigger an earlier merge or delay it beyond the merging interval suggested by the learning signal.

We define the forgetting signal based on the loss change of historical data during the training of new tasks.
In each iteration, a batch of historical data is sampled from the memory buffer and combined with the current task's batch. The historical data is used only for loss computation, excluding it from gradient updates. 
The forgetting signal is activated if the loss on historical tasks exceeds a predefined threshold, which is 
calculated using the average loss over the first $2/3 \times S_{b+1}$ steps, scaled by an adjustment factor $\gamma_{forget}$ to produce the threshold $\delta_{b+1}$.
If the historical loss exceeds this threshold during subsequent training, the forgetting signal is triggered and the activation count is incremented as: $\mathcal{F}(b+1) = \mathcal{F}(b+1) + 1$.
% We then monitor in real time whether the loss on historical tasks exceeds a predefined threshold to determine if the forgetting signal should be activated. The threshold is set as follows: assume the model has just completed the $b$-th merge, and the next merging interval given by the learning signal is $S_{b+1}$. The average loss on historical data is calculated using the first $2/3 \times S_{b+1}$ training steps, and this average is scaled by a adjustment factor $\gamma_{forget}$ to produce the threshold $\delta_{b+1}$ for the $(b+1)$-th merge. If the historical loss exceeds this threshold during subsequent training, the forgetting signal is activated and the count of activations is incremented as: $\mathcal{F}(b+1) = \mathcal{F}(b+1) + 1$.

\paragraph{Overall Workflow of the Merge Controller} If the forgetting signal is activated multiple times (e.g., $\mathcal{F}(b+1) \geq \mathcal{F}_{max}$) before the scheduled merging interval $S_{b+1}$, an early merge is triggered to prevent further forgetting, i.e., the actual merging interval $S^\prime_{b+1} < S_{b+1}$. 
Conversely, if the model reaches the predefined merging interval $S_{b+1}$ without the forgetting signal being activated, the merge can be deferred to allow the model to continue focusing on learning new knowledge. The merge will be triggered either when the forgetting signal is activated ($S^\prime_{b+1} > S_{b+1}$) or when the iteration count reaches the upper limit ($S^\prime_{b+1} = 2 * S_{b+1}$).

%In summary, our merge controller simultaneously considers the changes in both signals, maximizing new knowledge learning while minimizing the forgetting of old knowledge, resulting in a balanced strategy for dynamic control.
%In summary, our merge controller jointly considers both learning and forgetting signals to maximize the acquisition of new knowledge while minimizing the forgetting of prior knowledge, resulting in a balanced and dynamically adaptive merging strategy.

In summary, our controller dynamically balances both learning and forgetting signals to optimize new knowledge acquisition while minimizing forgetting, resulting in an adaptive merging strategy.

\subsection{Rehearsal-based Knowledge Fusion}

When the merge controller initiates a merge, the knowledge fusion module performs the actual task knowledge fusion.
Assume the $b$-th merge occurs at the $j$-th training iteration. The parameter change representing new knowledge is defined as:
\begin{equation}
\tau_{\text{new}_b} = \theta_j - \theta_{j - S^\prime_b}
\end{equation}

Next, we fine-tune $\theta_j$ on memory data for $S^\prime_b/2$ steps, resulting in an updated model state $\theta_{j(M)}$.
The task vector for historical knowledge is then:
\begin{equation}
\tau_{\text{past}_b} = \theta_{j(M)} - \theta_{j}
\end{equation}

The final model parameters are updated by fusing both task vectors with learnable weights:
%The knowledge fusion strategy combines $\tau_{\text{new}}$ and $\tau_{\text{his}}$ with weighted fusion into the model before merging, as follows:
\begin{equation}
\hat{\theta}_j = \theta_{j - S^\prime_b} + \alpha_1 \cdot \tau_{\text{new}_b} + \alpha_2 \cdot \tau_{\text{past}_b}
\end{equation}
where $\alpha_1$ and $\alpha_2$ are the fusion weights, computed as follows:
\begin{itemize}[leftmargin=*,itemsep=2pt,topsep=0pt,parsep=0pt]
\item 
For $\tau_{\text{new}}$, we assess the proportion of the upward trend in the learning signal's sliding window, $\mathcal{P}_{new} = L_{up} / L_w$, indicating the model’s active learning of new knowledge.

\item 
For $\tau_{\text{past}}$, we compute the ratio of the forgetting signal's activation count $\mathcal{F}(b)$ to the maximum threshold $\mathcal{F}_{\text{max}}$, $\mathcal{P}_{past} = \mathcal{F}(b) / \mathcal{F}_{\text{max}}$, suggesting the extent of historical knowledge forgetting.

\end{itemize}

The fusion weights are then normalized as:
\begin{equation}
\alpha_1 = \frac{\mathcal{P}_{new} }{\mathcal{P}_{new} + \mathcal{P}_{past} }, \quad \alpha_2 = \frac{\mathcal{P}_{past} }{\mathcal{P}_{new} + \mathcal{P}_{past} }
\end{equation}
%To ensure fairness in comparison, we set the length of the sliding window for the learning signal and $f_{\text{max}}$ for the forgetting signal to 3.
After the fusion is completed, training continues from the updated model state $\hat{\theta}_j$. 
%The complete algorithm is provided in Algorithm X.

\section{Experiments and Analysis}\label{sec:exp}
\paragraph{Dataset} We adopt the experimental setup from \citet{du2024unlocking}, using three CL benchmark datasets:
(i) \textbf{Standard CL Benchmark}, which consists of five text classification tasks from \citet{zhang2015character}.
(ii) \textbf{Long Sequence Benchmark}, a more challenging evaluation scenario comprising 15 tasks \cite{razdaibiedina2023progressive}.
(iii) \textbf{SuperNI Benchmark}~\cite{wang2022super}, a comprehensive benchmark for text generation, designed to evaluate 15 NLP tasks.
Following \citet{wang2023orthogonal}, we sample 1000 instances for training on each task and reserve 500 per class for testing. Different task sequences are evaluated for each benchmark, with detailed descriptions provided in Appendix \ref{sec:dataset}.

\newcommand{\tabincell}[2]{\begin{tabular}{@{}#1@{}}#2\end{tabular}}
\begin{table*}[t]

\centering
\scalebox{0.85}{
\begin{tabular}{l|ccc|ccc|ccc}
\toprule
\multirow{2}*{\tabincell{c}{ }}  & \multicolumn{3}{c|}{Standard CL } & \multicolumn{3}{c|}{Long Sequence }  & \multicolumn{3}{c}{SuperNI } \\

%& OP & FWT & BWT & OP & FWT  & BWT & OP & FWT$\uparrow$ & BWT$\uparrow$ \\

& OP$\uparrow$ & FWT$\uparrow$ & BWT$\uparrow$ & OP$\uparrow$ & FWT$\uparrow$ &   BWT$\uparrow$ & OP$\uparrow$ & FWT$\uparrow$ & BWT$\uparrow$ \\
  
\midrule
\rule{0pt}{4pt}SeqLoRA  & 43.7 & -9.1 & -50.4 & 11.6 & -10.8 & -73.4  & 6.4 & -13.6 & -31.0\\
\rule{0pt}{8pt}IncLoRA  & 66.4 & -8.7 &  -20.0 & 61.2 & -11.1 & -26.7 & 8.2 & -15.1 & -27.4 \\
\rule{0pt}{8pt}LoRAReplay  & 68.8 & -9.0 &  -11.7 & 70.9 & -11.3 &  -15.4  & 35.4 & -12.4 & -15.8\\
\rule{0pt}{8pt}EWC$^*$~\cite{kirkpatrick2017overcoming} & 50.3 & - &  - & 45.1 & - & - & 35.7 & - & - \\
\rule{0pt}{8pt}L2P$^*$~\cite{wang2022learning}  & 60.7 & - &  - & 56.1 & 
 1.36 &   -16.3 & 12.7 & -19.1 & -8.0\\
\rule{0pt}{8pt}LFPT5$^*$~\cite{qin2021lfpt5}  & 72.7 & - & - & 69.2 & -2.5 & -12.8 & 34.4 & -0.5 & -14.5\\
\rule{0pt}{8pt}MoELoRA$^*$~\cite{luo2024moelora} & 54.1 & -6.2 &  -7.7 & 27.6&  -8.6  & -13.2 & 21.8 & -7.2 & -19.0 \\
\rule{0pt}{8pt}O-LoRA$^*$~\cite{wang2023orthogonal} & 75.8 & -5.9  & -3.8  & 69.6 & -8.2  & -4.1 & 25.9 & -0.1 & -24.6 \\
\rule{0pt}{8pt}TaSL~\cite{feng2024tasl} & 76.3 & -5.4 &  -4.0 & 74.4&  -7.9 & -5.3  & 38.9 &  -1.2 & -10.8  \\
\rule{0pt}{8pt}MIGU$^*$~\cite{du2024unlocking} & 76.6 & - &  - & 76.5 & - & -  & - & - & - \\
\rule{0pt}{8pt}VR-MCL~\cite{wu2024meta}  & 76.0 &  -4.6&   -3.7 & 74.8 & -6.0 & -4.9 & 41.1 & 0.2 & -9.3\\
\rule{0pt}{8pt} SAPT-LoRA~\cite{zhao2024sapt} & - & - & - & 76.6 & -5.1 & -3.7 & 41.7 & 1.9 & -6.7  \\

\rule{0pt}{8pt}Recurrent-KIF$^*$~\cite{feng2025recurrent} &\textbf{78.4}  & -3.1 & -2.8 & 77.8 & -4.6  &  -3.6 & 43.3 & 0.4  & -8.4 \\
\rowcolor[gray]{0.9}
\rule{0pt}{8pt}\textbf{{\ouralg} (ours)} & 78.1 &  \textbf{-1.5} & \textbf{-0.4} & \textbf{77.9} & \textbf{-2.3} &  \textbf{-1.8}  &  \textbf{44.3} &  \textbf{2.2} &  \textbf{-4.0}\\

\midrule

\rule{0pt}{8pt} MTL &  80.3 & -& - & 81.8& - & - & 50.7 & - & - \\

\bottomrule
\end{tabular}}
\caption{Overall results on three CL benchmarks using the T5-large model. We report Overall Performance (OP), Forward Transfer (FWT), and Backward Transfer (BWT) after training on the final task. All results are averaged over different task orders. Methods marked with $*$ are copied from previous papers. The last row represents upper bound performance.
%MTL represents multi-task learning results, which serve as an upper bound.
%Our method, {\ouralg}, outperforms the previous best method, MIGU, with an average improvement of 1.5\% in OP and a xx\% increase in BWT.
}
\label{tbl:result}
\end{table*}

\paragraph{Metrics}
Let $a_{i,j}$ denote the testing performance on task $\mathcal{T}_i$ after training on task $\mathcal{T}_j$, and \( a_{0,t} \) refers to the performance of training task \( t \) individually. We evaluate the overall performance (OP) \cite{chaudhry2018riemannian}, backward transfer (BWT) \cite{ke2022continual}, and forward transfer (FWT) \cite{lopez2017gradient} after training on the final task:
\begin{equation}
    \mathbf{OP} =\frac{1}{K} \sum_{i=1}^{K} a_{i, K}
\end{equation}
\begin{equation}
    \mathbf{BWT} = \frac{1}{K-1} \sum\limits_{i=1}^{K-1} (a_{i, K}-a_{i, i})
\end{equation}
\begin{equation}
\text{FWT} = \frac{1}{K} \sum_{i=1}^{K} (a_{i,i} - a_{0,i}),
\end{equation}

\paragraph{Baselines}
We compare {\ouralg} against various advanced methods, %including recent CL approaches, 
as well as both single-round and multi-round model merging methods. All methods are implemented using the LoRA framework for fairness. 
%Descriptions of these methods are provided in Appendix \ref{sec:baselines}.
(1) \textit{\textbf{SeqLoRA}:} LoRA parameters are trained on a task sequence without regularization or sample replay.
(2) \textit{\textbf{IncLoRA}:} incremental learning of LoRA parameters without regularization or sample replay.
(3) \textit{\textbf{LoRAReplay}:} LoRA fine-tuning with a memory buffer.
(4) \textit{\textbf{EWC \cite{kirkpatrick2017overcoming}}:} finetune LoRA with a regularization loss to prevent interference with previous tasks.
(5) \textit{\textbf{L2P \cite{wang2022learning}}:} dynamically selects and updates prompts from a pool on an instance-by-instance basis.
(6) \textit{\textbf{LFPT5 \cite{qin2021lfpt5}}:} learns a soft prompt that solves tasks and generates training samples for replay.
(7) \textit{\textbf{O-LoRA \cite{wang2023orthogonal}}:} extends IncLoRA to
learn different LoRAs in orthogonal subspaces.
(8) \textit{\textbf{MoELoRA \cite{luo2024moelora}}:} a vanilla MoE with LoRA number equals to the task number.
(9) \textit{\textbf{SAPT \cite{zhao2024sapt}}:} uses pseudo samples and a shared attention framework to align PEFT block learning and selection.
(10) \textit{\textbf{MIGU \cite{du2024unlocking}}:} updates important parameters based on gradient magnitude.
(11) \textit{\textbf{TaSL \cite{feng2024tasl}}:} a single-round model merging method based on parameter importance.
(12) \textit{\textbf{VR-MCL \cite{wu2024meta}}:} dynamically updates the distribution of parameter importance through memory replay.
(13) \textit{\textbf{Recurrent-KIF \cite{feng2025recurrent}}:} a multi-round model merging method based on fixed merging intervals.
%Additionally, multi-task learning on LoRA is reported as \textit{\textbf{MTL}}, serving as the upper-bound.
Additionally, multi-task learning, referred to as \textit{\textbf{MTL}}, serves as the upper bound.

\paragraph{Training Details}
We evaluate {\ouralg} using different backbone models, including T5-large  \cite{raffel2020exploring}, Qwen3 1.7B \cite{qwen3}, LLaMA2-7B \cite{touvron2023llama}, and LLaMA2-13B.
For the learning signal, the initial merging interval $S_{init}$ is set to 8, and the sliding window size $L_w$ is set to 3. The minimum and maximum merging intervals, $S_{min}$ and $S_{max}$, are set to 2 and 128, respectively. The adjustment factors $\gamma_{learn}^{+}$ and $\gamma_{learn}^{-}$ are selected based on the current merging interval: if \(S > 64\), we set \(\gamma_{\text{learn}}^+ = 1.5\) and \(\gamma_{\text{learn}}^- = 2\); otherwise, we set \(\gamma_{\text{learn}}^+ = 2\) and \(\gamma_{\text{learn}}^- = 1.5\).
For the forgetting signal, the threshold scaling factor \(\gamma_{\text{forget}}\) is set to 2, and the maximum number of allowed activations before forcing an early merge, \(\mathcal{F}_{\text{max}}\), is set to 3.
Following \citet{feng2025recurrent}, 2\% of the original training set is used for replay samples.
%It is worth noting that we used the same hyperparameters across different datasets and backbones, demonstrating the generalizability of our method without requiring extensive hyperparameter tuning for each specific setting.
All experiments are averaged over 3 runs. More details are in Appendix \ref{sec:details}.

\subsection{Main Results}

The overall CL results using the same T5-large backbone are summarized in Table \ref{tbl:result}.

\paragraph{Our Training Trajectory-based {\ouralg} Method Effectively Addresses Both CF and KT Challenges.}
Compared to traditional CL methods (LoRAReplay, EWC) and model merging approaches (O-LoRA, MoELoRA, TaSL), {\ouralg} outperforms them in both CF (increasing OP from 59.5\% to 66.8\% compared to O-LoRA) and KT (improving BWT from -6.0\% to -2.1\% compared to VR-MCL). 
%Compared to traditional CL methods (LoRAReplay, EWC) and model merge-based methods (O-LoRA, MoELoRA, TaSL), {\ouralg} outperforms them in both CF (increasing average OP from 59.5\% to 66.8\% compared to O-LoRA) and KT (increasing average BWT from -6.0\% to -2.1\% compared to VR-MCL). SAPT achieves the highest performance by leveraging generative replay-based data augmentation, even surpassing MTL results. Notably, our method demonstrates a BWT value on the long sequence benchmark close to SAPT's value (-1.8\% vs -1.3\%), and exceeds SAPT's FWT on the SuperNI benchmark (improving from 1.9\% to 2.2\%).
Moreover, {\ouralg} outperforms the state-of-the-art Recurrent-KIF, also based on multi-round merging, with significant improvements in both FWT (2.0\%, from -2.5\% to -0.5\%) and BWT (2.9\%, from -4.9\% to -2.0\%). These results show that {\ouralg} effectively balances preserving prior knowledge and excelling in new tasks.
%Moreover, {\ouralg} outperforms the state-of-the-art CL method, Recurrent-KIF, which is also based on a multi-round merging approach. While maintaining a modest increase in the OP metric, {\ouralg} achieves significant improvements in both FWT and BWT. The average improvements are 2.0\% (from -2.5\% to -0.5\%) and 2.9\% (from -4.9\% to -2.0\%), respectively. These results demonstrate that {\ouralg} effectively strikes a balance between preserving prior knowledge and excelling in new tasks.

\begin{figure}[t]
  \centering
  \includegraphics[width=1\linewidth]{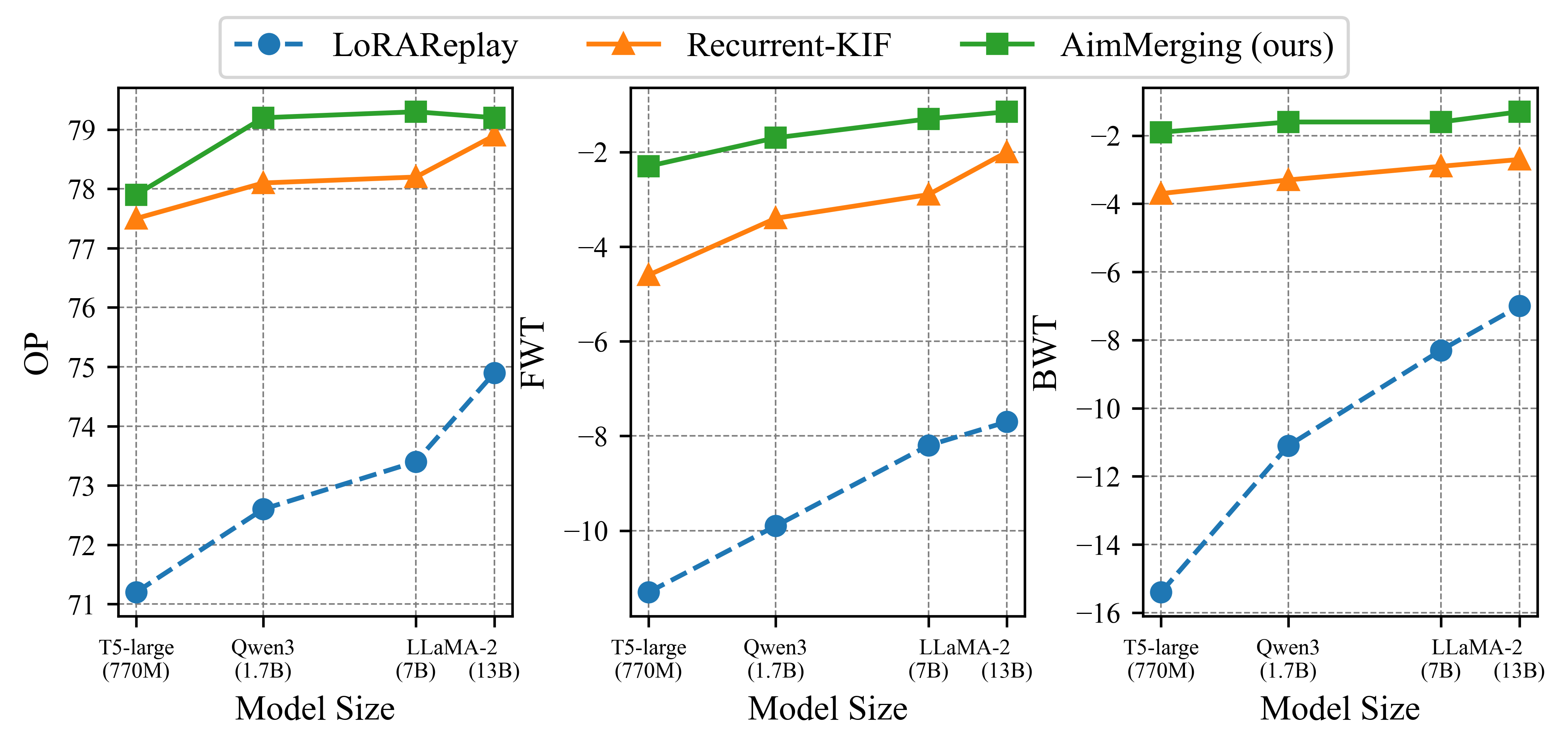}
  \caption{Performance of {\ouralg} with different backbones on the Long Sequence Benchmark.}
  \label{fig:different_size}
\end{figure}

\paragraph{{\ouralg} Demonstrates Consistent Superiority and Generalization Across Various Backbones.}
We validated the robustness of {\ouralg} using backbones ranging from 770M to 13B parameters, as shown in Figure \ref{fig:different_size}. Across all sizes, {\ouralg} consistently outperforms baseline models. Notably, with the LLaMA2-7B backbone, {\ouralg} improves FWT from 78.2\% to 79.3\% and BWT from -2.9\% to -1.6\% compared to Recurrent-KIF, demonstrating its strong generalization ability across different model scales.
%To further validate the robustness of {\ouralg}, we conducted experiments using a range of parameter-sized backbones, as shown in Figure \ref{fig:different_size}. Across all backbone sizes, from 770M to 13B parameters, {\ouralg} consistently outperforms all baseline models. Notably, using the LLaMA2-7B backbone, {\ouralg} improves the FWT metric from 78.2\% to 79.3\% and increases BWT from -2.9\% to -1.6\% compared to Recurrent-KIF. These results further highlight the strong generalization capability of {\ouralg} across different model scales.

\begin{figure}[t]
  \centering
  \includegraphics[width=0.9\linewidth]{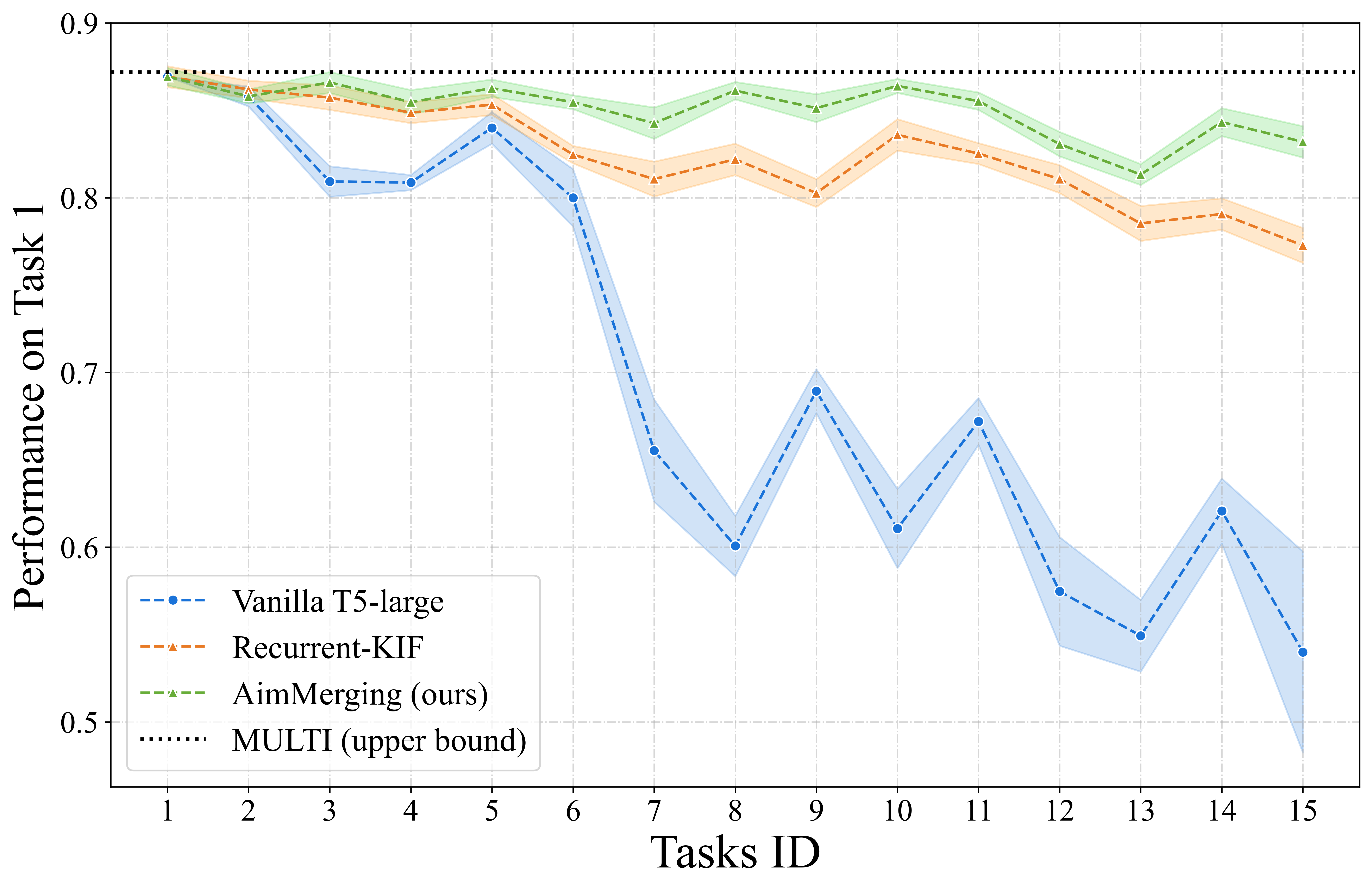}
  \caption{Performance trajectory of Task 1 on the longsequence benchmark during the CL process.
  }
  \label{fig:forgetting}
\end{figure}

\paragraph{The Adaptive Iterative Merging Framework Enables Effective Knowledge Retention.}
Figure \ref{fig:forgetting} illustrates the performance of the initial task after training on subsequent tasks. {\ouralg} significantly reduces catastrophic forgetting, with only a 4\% performance drop after training on the final task. In contrast, vanilla replay shows a 32\% drop, and Recurrent-KIF shows a 10\% decline. These results underscore our model's strong backward knowledge transfer capability.
%Figure \ref{fig:forgetting} illustrates the performance of the initial task after training on subsequent tasks. The results demonstrate that {\ouralg} results in a significantly slower forgetting rate, with an average performance decrease of only 4\% after training on the final task. In contrast, vanilla replay shows a substantial performance drop of 32\%, while Recurrent-KIF experiences a decline of 10\%. These results highlight our model's strong backward knowledge transfer capability.

% Figure \ref{fig:radial_plot} presents the performance of various methods across all historical tasks after completing the final task. It shows that {\ouralg} optimally restores performance on previous tasks, with notable improvements on tasks xx and xx. Remarkably, on tasks xx and xx, {\ouralg} performs similarly to multi-task learning results. These findings highlight that {\ouralg} strikes a robust balance between preserving prior knowledge and excelling in new tasks.

% \begin{figure}[t]
%   \centering
%   \includegraphics[width=1\linewidth]{imgs/Radial.pdf}
%   \caption{\textbf{Impact of Catastrophic Forgetting in Continual Learning.}
%   After fine-tuning on the final task (in orange), {\ouralg} demonstrates superior resistance to performance decline on previously learned tasks (in blue), outperforming baseline methods.
%   }
%   \label{fig:radial_plot}
% \end{figure}

\begin{table}[t]
\centering
\scalebox{0.9}{
\begin{tabular}{lccc}
\toprule
Method & OP &  FWT & BWT\\
\midrule
\rowcolor[gray]{1}
\rule{0pt}{6pt} {\ouralg}  & \textbf{45.1} & \textbf{1.3}  & -2.2 \\
\midrule
\rule{0pt}{8pt}  - LS  & 43.9 & 0.5 & -3.4 \\
\rule{0pt}{8pt}  - FS &  44.3 & 0.8 & -3.9 \\
\rule{0pt}{8pt} + MGM &  44.2 & 0.7 & -3.7 \\
\rule{0pt}{8pt} + IFM & 44.9 & 1.2 & \textbf{-2.1}   \\

\bottomrule
\end{tabular}}
\caption{Ablation study. ``- LS'', ``- FS'' refer to the removal of the learning signal and forgetting signal in our merge controller, respectively. ``+ MGM'' and ``+ IFM'' represent replacing the merging weights with manually set global merging weights and parameter importance-based fine-grained merging weights, respectively.
}
\label{tbl:ablation}
\end{table}

\subsection{Ablation Study}
We perform ablation studies to evaluate the effectiveness of the two key techniques in {\ouralg}. Results for task order 1 on the SuperNI Benchmark are shown in Table \ref{tbl:ablation}. Additional experiments, such as time complexity analysis and memory size impact, are provided in Appendix~\ref{sec:hyper}.

\paragraph{Effect of Training Trajectory-guided Merge Controller.}  
To validate the contribution of the learning signal and forgetting signal to the merge controller’s decision-making, we remove the learning signal (``- LS'') and forgetting signal (``- FS'') individually. When only the learning signal is used, merging occurs whenever the model’s iteration reaches the pre-defined interval $S$. When only the forgetting signal is used, merging occurs when the loss of historical tasks exceeds the threshold.
The performance decline in Table \ref{tbl:ablation} highlights the necessity of both signals. 
Using only one signal leads to focusing on either new knowledge learning or historical knowledge retention, while both signals allow better balance.
%Relying on only one supervisory signal causes the model to focus either on learning new knowledge or retaining old knowledge. However, by simultaneously monitoring both signals, we can better balance the performance of both new and historical knowledge.

\paragraph{Effect of Rehearsal-based Knowledge Fusion Module.}
We replace the weight calculation method in our fusion mechanism with two alternatives:
(i) Manually set global merging weights (via grid search).
(ii) Parameter importance-based fine-grained merging weights (following \citet{feng2025recurrent}).
Our results show that weights based on the learning and forgetting signals outperform manually set weights, improving three evaluation metrics by 0.9\%, 0.6\%, and 1.5\%. Compared to parameter importance-based weights, our method shows slightly better performance, demonstrating that the learning and forgetting signals effectively capture the relevance of task vectors for knowledge updating and retention. Moreover, our approach is more efficient, avoiding the computational overhead of calculating and storing parameter importance.
%The results show that the weights set based on the learning and forgetting signal outperformed the manually set weights, with improvements of 0.9\%, 0.6\%, and 1.5\% on three evaluation metrics. Compared to parameter importance-based weights, our method demonstrates slightly better performance. This suggests that the learning and forgetting signals effectively capture the relevance of new and historical task vectors in terms of knowledge updating and retention. Moreover, our method is more efficient, as it avoids the computational and storage overhead associated with calculating and storing parameter importance.

\subsection{Effect of Adding LoRA at Different Positions in the Model}
We further investigate the impact of adding LoRA to different positions within the Transformer block. A typical Transformer block consists of the query, key, and value (QKV) linear layers, the output linear layer (O) in the multi-head attention module, and the two linear layers in the feedforward network (FFN). Our analysis, presented in Table \ref{tbl:ablation_lora}, demonstrates that applying LoRA to all of these linear layers results in the best overall performance. 

\begin{table}[t]
\centering
\scalebox{1}{
\begin{tabular}{lrrr}
\toprule
LoRA Target Modules &   OP  & FWT & BWT \\
\midrule
\rule{0pt}{4pt} Attention Q V  & 45.1 & 1.3  & -2.2  \\

\rule{0pt}{8pt} Attention Q K V O & 45.3  & 1.4 & -2.3 	   \\
\rule{0pt}{8pt} FFN  & 49.8   &  1.0 & -1.9 	   \\
\rule{0pt}{8pt} Attention All + FFN & 45.7  &2.1 & 	-2.0    \\

\bottomrule
\end{tabular}}
\caption{Ablation study on LoRA target modules, using T5-large as the backbone.}
\label{tbl:ablation_lora}
\end{table}

\subsection{Visualization}
We visualize two key aspects of our method's effectiveness.
% present two key visualizations to analyze the effectiveness of our proposed methods, 
Full results for all tasks are provided in Appendix \ref{appendix:vis} (Figure \ref{fig:Vis_controller_superni} - \ref{fig:Vis_forget_stand}).
%(Figure \ref{fig:Vis_controller_superni}, \ref{fig:Vis_controller_long}, \ref{fig:Vis_controller_stand}, \ref{fig:Vis_forget_superni}, \ref{fig:Vis_forget_long}, and \ref{fig:Vis_forget_stand}).

\begin{figure}[t]
  \centering
  \includegraphics[width=1\linewidth]{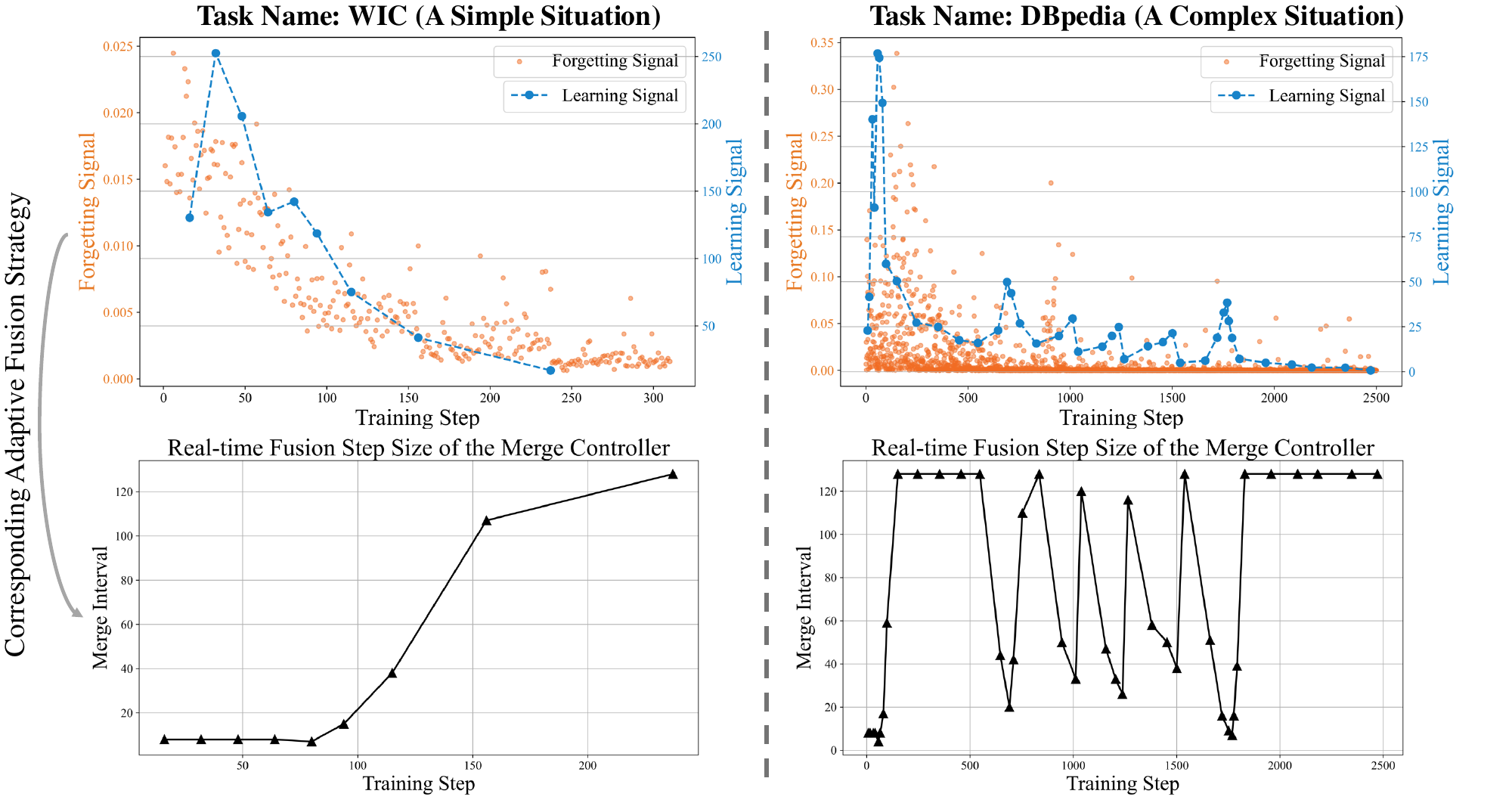}
  \caption{Visualizing merge controller behavior based on dynamic changes in learning and forgetting signals.}
  \label{fig:Vis_controller}
\end{figure}

\paragraph{Visualizing How the Merge Controller Adjusts Merging Timing and Step Size Based on the Learning and Forgetting Signals.}
%As shown in Figure \ref{fig:Vis_controller}, in simpler training scenarios, the merge controller gradually increases the merging interval. Early on, increasing merging frequency prevents catastrophic forgetting, while in later stages, reducing it avoids redundant merges. In more complex scenarios (right of Figure \ref{fig:Vis_controller}), the merging frequency is dynamically adjusted based on training changes, such as during multiple rapid learning phases (indicated by peaks), prompting more frequent merges.
As shown on the left of Figure \ref{fig:Vis_controller}, for simpler training scenarios, the merge controller gradually increases the merging interval. In the early stages of training, when there is significant new knowledge update, increasing the merging frequency helps prevent forgetting of historical knowledge. While in later stages, reducing it avoids redundant merges.
In contrast, for more complex scenarios shown on the right, our method dynamically adjusts the merging frequency based on changes in the training state. For example, the model enters multiple rapid learning phases again during the middle of training (indicated by the peaks in the figure), prompting an increase in merging frequency.

% for the same task, the merge controller adaptively adjusts the merging interval and timing based on the varying states of the learning and forgetting signals during training. Furthermore, for different tasks, {\ouralg} adjusts the merging trajectory according to the specific learning state of each task.

\begin{figure*}[htbp]
  \centering
  \includegraphics[width=1\linewidth]{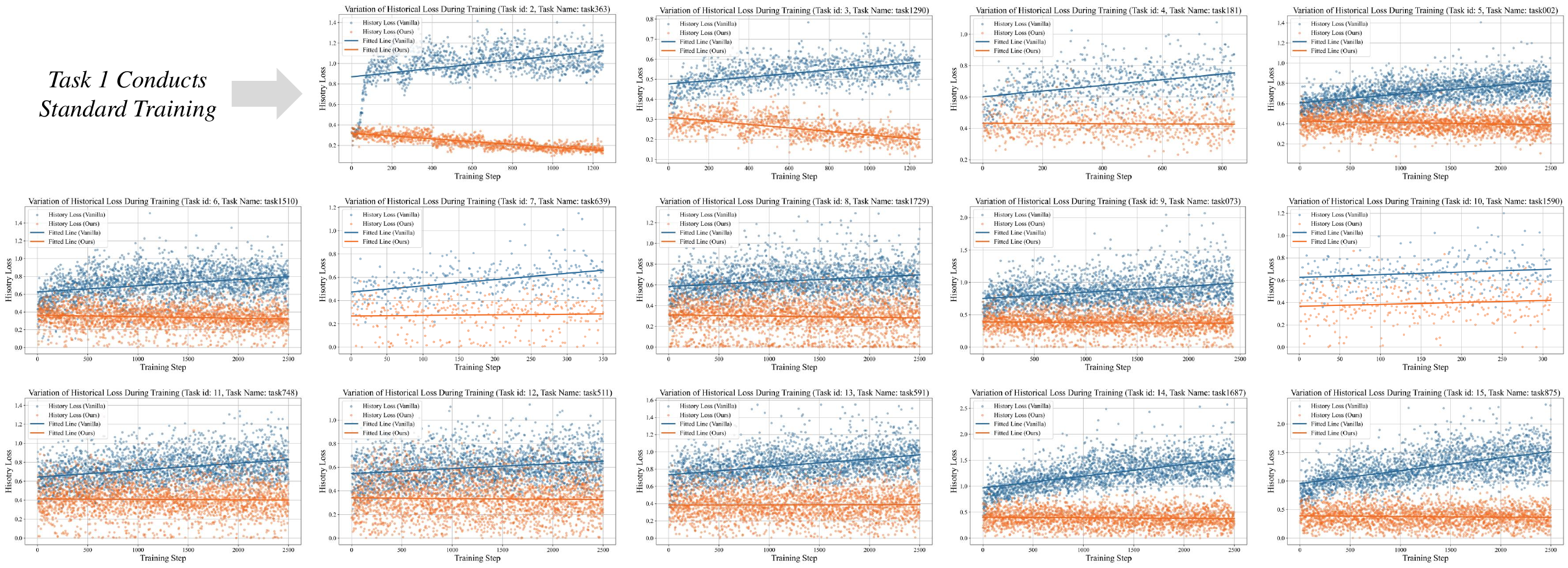}
  \caption{Visualization of the effect of {\ouralg} in alleviating catastrophic forgetting in the SuperNI benchmark.}
  \label{fig:Vis_forget}
\end{figure*}

\paragraph{Visualizing Adaptive Iterative Merging's Effect on Catastrophic Forgetting.}
Figure \ref{fig:Vis_forget} demonstrates the impact of our multi-round merging approach on historical knowledge forgetting. With vanilla LoRAReplay, the loss for historical tasks increases progressively, reflecting an escalating degree of forgetting. In contrast, our method effectively mitigates forgetting by selecting optimal merging points, enabling timely suppression before significant forgetting occurs. This results in a more stable or even decreasing overall loss trend, demonstrating the effectiveness of our approach in alleviating catastrophic forgetting.

%Figure \ref{fig:Vis_forget} demonstrates the impact of our multi-round merging approach on historical knowledge forgetting. With vanilla LoRAReplay, historical task loss increases steadily, showing growing forgetting. In contrast, our method mitigates this by selecting optimal merging points, stabilizing or even reducing the overall loss, effectively alleviating catastrophic forgetting.

% \begin{figure}[t]
%   \centering
%   \includegraphics[width=1\linewidth]{imgs/Vis_forget.png}
%   \caption{Visualization of the effect of {\ouralg} in alleviating catastrophic forgetting.}
%   \label{fig:Vis_forget}
% \end{figure}

\section{Related Work}
% \subsection{Continual Learning for LLMs}

Continual learning (CL) \cite{zhou2024continual} focuses on the development of algorithms that enable models to accumulate knowledge from non-stationary data. 
In the era of LLMs, model mixture-based methods that employ parameter-efficient fine-tuning (PEFT) have become the dominant approach \cite{huang2024mitigating, shi2024understanding, zhong2024lowrankinterconnectedadaptationlayers}, typically falling into two categories: model ensemble and model merging techniques.

Model ensemble methods allocate independent PEFT blocks to each task, effectively isolating task-specific parameters \cite{feng2023towards, pham2023continual, ke2023sub, shengyuan2023differentiable, li2024revisiting, he2024seekr, wang2024self}. For instance, O-LoRA \cite{wang2023orthogonal} enforces orthogonality among LoRA adapters, while SAPT \cite{zhao2024sapt} uses a selection module to combine task-specific blocks via task correlations.  
% Although these methods help preserve task-specific knowledge, they limit inter-task transfer and incur significant memory overhead, making them less scalable as the number of tasks increases \cite{zhang2025survey}.
Though effective for knowledge preservation, they hinder inter-task transfer and scale poorly due to growing memory overhead \cite{zhang2025survey}.

In contrast, model merging techniques combine multiple models into a single unified model \cite{cheng2024dam, alexandrov2024mitigating, ren2024analyzing}, addressing memory constraints. For example, global model merging approaches \cite{wortsman2022model, ilharco2023editing} perform a weighted fusion of models before and after training, often assuming that all model parameters contribute equally to each task.
Fine-grained approaches like \citet{feng2024tasl2} leverage parameter importance masks to enable neuron- or matrix-level fusion. 
% And \citet{feng2024tasl} proposed a fine-grained model merging method that utilizes parameter importance to set masks, enabling neuron-level or matrix-level model fusion. 
Recently, \citet{feng2025recurrent} introduced a multi-round merging paradigm for CL,
demonstrating that integrating merges during model iterations can significantly enhance model performance.
% and exhibit substantial potential for improvement. 
Yet key challenges persist: the optimal number, timing, and frequency of merges remain underexplored.
% However, the optimal number of merges, as well as the best timing and frequency of merges, remain open areas of exploration.
To address this, we propose {\ouralg}, a novel adaptive iterative framework that leverages learning and forgetting signals to dynamically monitor the model's state. 
By analyzing training trajectory, 
{\ouralg} optimizes merging strategies, advancing the efficiency and effectiveness of CL.
% {\ouralg} selects the optimal merging timing and frequency, further enhancing the effectiveness of CL.

\section{Conclusion}

In this paper, we introduce Adaptive Iterative Model Merging ({\ouralg}), a novel CL framework that enables dynamic monitoring of the training status by leveraging learning and forgetting signals extracted from the training trajectory. The framework consists of two key modules: the training trajectory-guided merge controller, which adaptively schedules the timing and frequency of model merging, and the rehearsal-based knowledge fusion module, which performs the global merging operation based on these signals. Extensive experiments validate the effectiveness of {\ouralg} in addressing the key challenges of continual learning.

\section*{Limitations}
% We acknowledge two limitations in this work.
% First, the core idea of our approach lies in selecting the appropriate timing for model merging based on the training trajectory. While our method effectively considers both the model's learning state for new tasks and the forgetting of historical tasks, there remains potential for improvement in how we measure these aspects. Currently, we rely on changes in model parameters and the loss of historical tasks as indicators. However, it is still open to exploration whether other metrics, such as gradients or other relevant information, could provide more effective measurements.

% Second, the strategies we use to adjust the merging process, such as the merging interval and threshold selection, involve a degree of subjective design. Future work could focus on developing fully automated methods to optimize these adjustments, reducing reliance on manual tuning.

We acknowledge two limitations in our work. 
First, while our approach selectively determines model merging timing by monitoring parameter changes and historical task loss, it remains an open question whether alternative metrics such as gradient information or other indicators could more effectively capture learning states and forgetting phenomena. Second, current merging strategies involve semi-heuristic design choices regarding intervals and thresholds. Future research could focus on developing fully automated optimization methods that minimize the need for manual parameter tuning.

% \section*{Ethics Statement}

% \section*{Acknowledgements}

\bibliography{emnlp2023}
\bibliographystyle{acl_natbib}

\appendix

\section{Visualization}
\label{appendix:vis}

\begin{figure*}[htbp]
  \centering
  \includegraphics[width=1\linewidth]{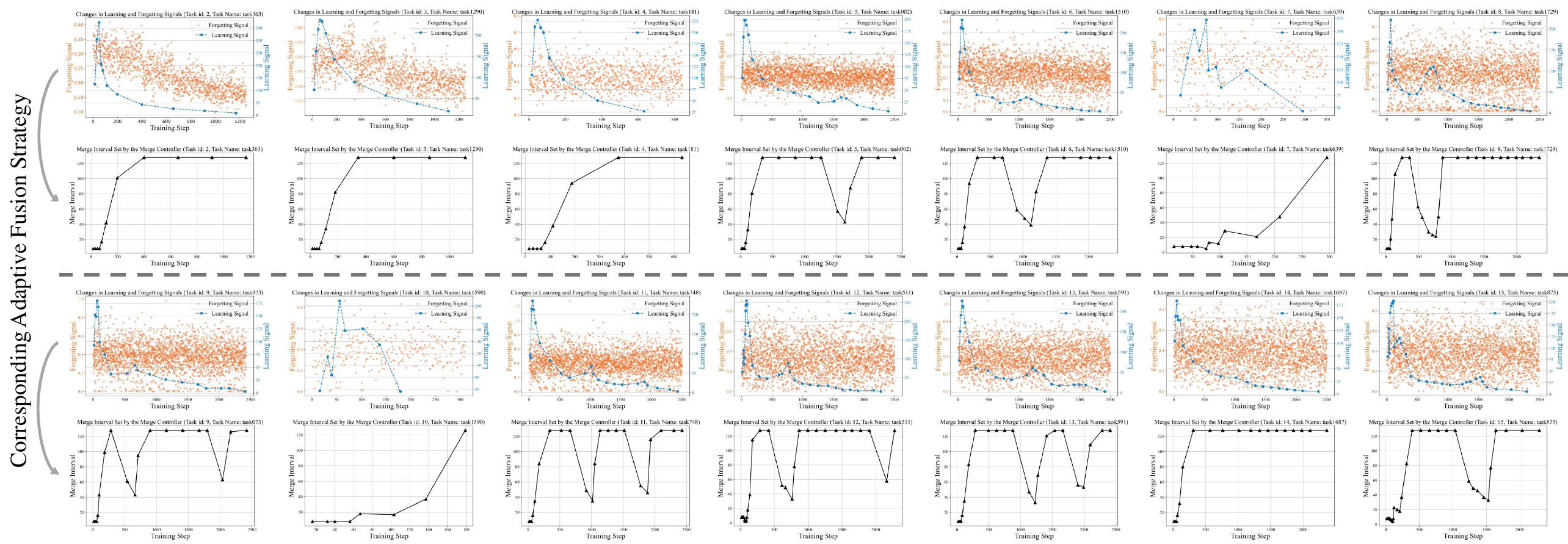}
  \caption{Visualizing the behavior of the merge controller based on dynamic changes in learning and forgetting signals in the SuperNI benchmark.}
  \label{fig:Vis_controller_superni}
\end{figure*}

\begin{figure*}[htbp]
  \centering
  \includegraphics[width=1\linewidth]{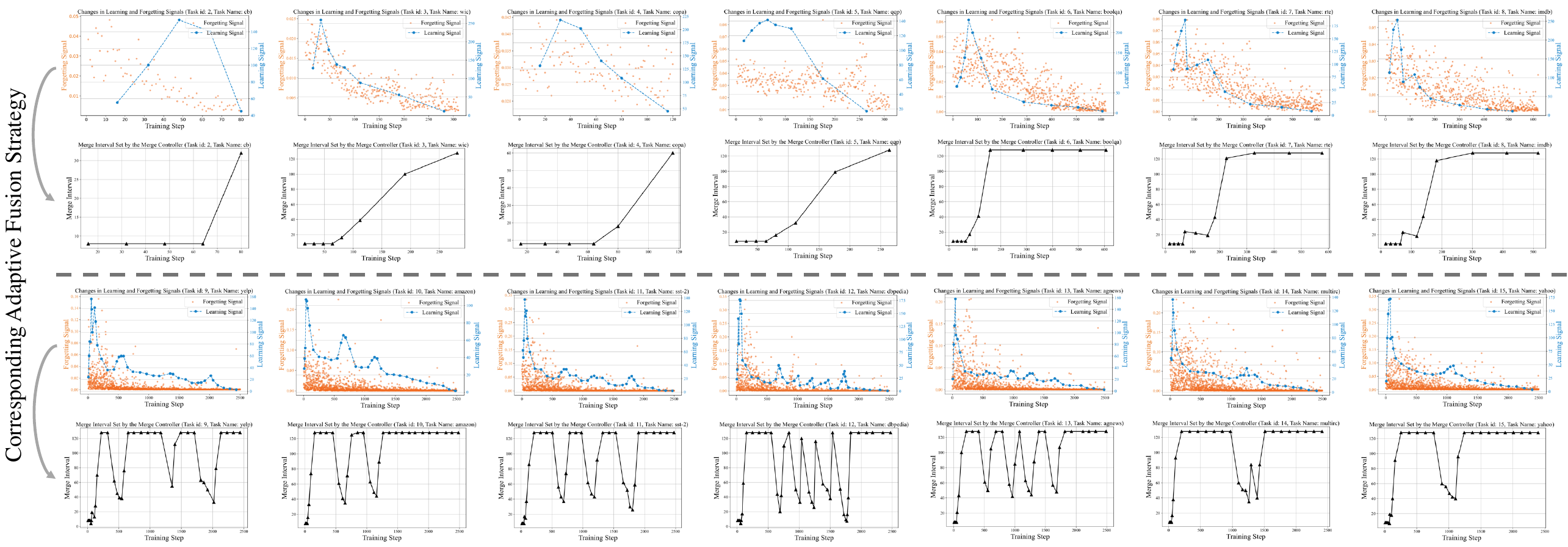}
  \caption{Visualizing the behavior of the merge controller based on dynamic changes in learning and forgetting signals in the Long Sequence benchmark.}
  \label{fig:Vis_controller_long}
\end{figure*}

\begin{figure*}[htbp]
  \centering
  \includegraphics[width=1\linewidth]{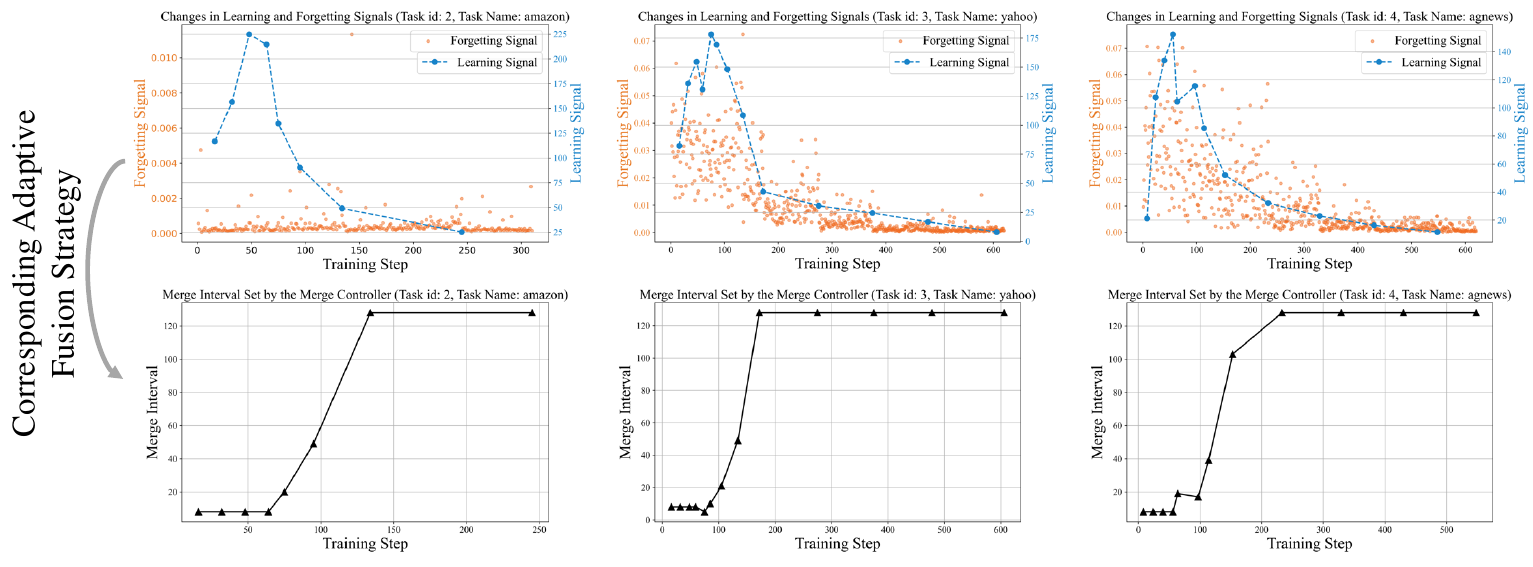}
  \caption{Visualizing the behavior of the merge controller based on dynamic changes in learning and forgetting signals in the Standard CL benchmark.}
  \label{fig:Vis_controller_stand}
\end{figure*}

\begin{figure*}[htbp]
  \centering
  \includegraphics[width=1\linewidth]{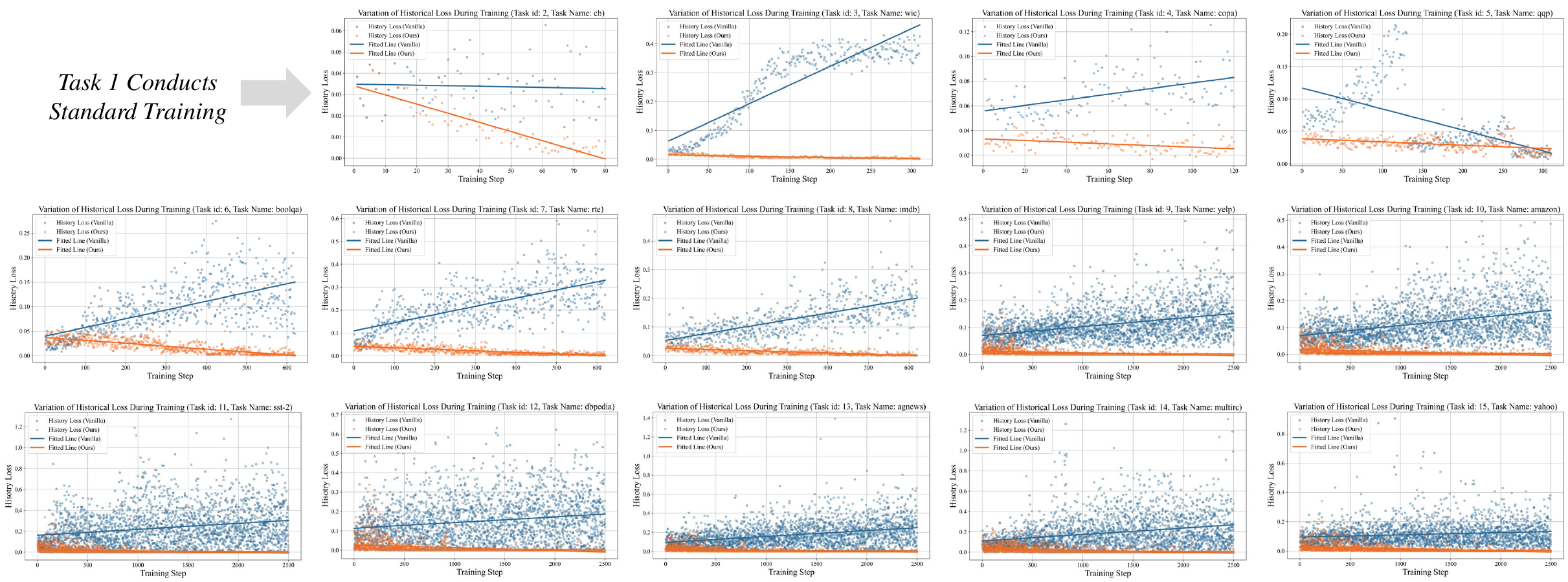}
  \caption{Visualization of the effect of {\ouralg} in alleviating catastrophic forgetting in the Longsequence benchmark.}
  \label{fig:Vis_forget_long}
\end{figure*}

\begin{figure*}[htbp]
  \centering
  \includegraphics[width=1\linewidth]{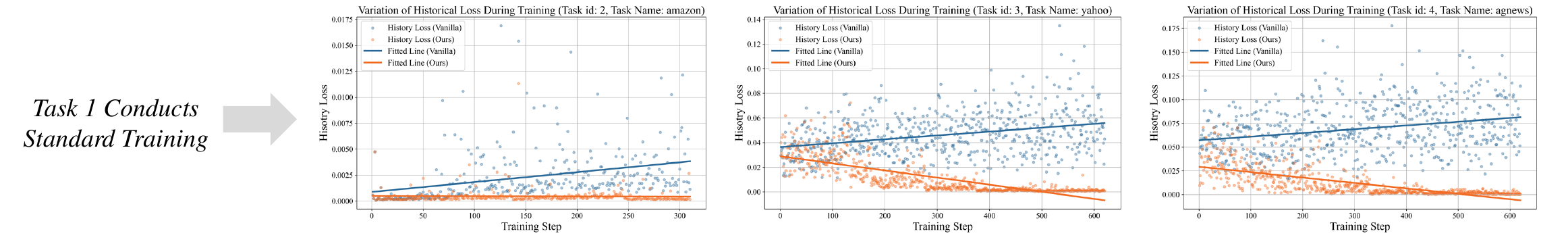}
  \caption{Visualization of the effect of {\ouralg} in alleviating catastrophic forgetting in the Standard CL benchmark.}
  \label{fig:Vis_forget_stand}
\end{figure*}

Here, we present the performance of our method across all datasets and tasks.
Figures \ref{fig:Vis_controller_superni}, \ref{fig:Vis_controller_long}, and \ref{fig:Vis_controller_stand} illustrate how our merge controller adjusts the merging strategy for all tasks in the SuperNI, LongSequence, and Standard benchmarks, respectively.
Figures \ref{fig:Vis_forget_long} and \ref{fig:Vis_forget_stand} also demonstrate the effectiveness of our method in alleviating catastrophic forgetting across all tasks.

% \begin{figure*}[htbp]
%   \centering
%   \includegraphics[width=1\linewidth]{imgs/vis_forgetting_superni.pdf}
%   \caption{Visualization of the effect of {\ouralg} in alleviating catastrophic forgetting in the SuperNI benchmark.}
%   \label{fig:Vis_forget_superni}
% \end{figure*}

\section{Additional Results}
\label{sec:hyper}
\subsection{Effect of the Memory Size}
We examine the effect of varying memory size on the performance of LoRAReplay and {\ouralg}. By adjusting the memory size per task \(|M|\) to {2\%, 5\%, 10\%, 50\%}, the results are presented in Table \ref{tbl:ablation_memory}. As anticipated, increasing the memory size generally enhances the performance of all methods. However, {\ouralg} utilizes iterative knowledge fusion mechanism to effectively retain historical knowledge, resulting in superior performance compared to LoRAReplay.

\begin{table}[h]
\centering
\scalebox{1}{
\begin{tabular}{lcccc}
\toprule
\multirow{2}*{\tabincell{c}{ }} & \multicolumn{4}{c}{Memory Size}\\
\cmidrule(lr){2-5}
 & 2\% & 5\% & 10\% & 50\%\\
\midrule
\rule{0pt}{6pt} LoRAReplay & 71.2 & 72.4& 73.8 & 76.1   \\
%\rule{0pt}{8pt} Recurrent-KIF & 77.8 & 78.7  & 79.8 & 80.5 \\
\rule{0pt}{8pt} {\ouralg} & 77.9 & 78.9  & 78.3 & 80.9 \\

\bottomrule
\end{tabular}}
\caption{Ablation study on memory size, using T5-large as the backbone.
%Average performance (AP) of Replay, TaSL-M, and TasLoRA-M across different memory sizes on the Long Sequence Benchmark.
}
\label{tbl:ablation_memory}
\end{table}

% \subsection{Effect of Adding LoRA at Different Positions in the Model}
% We further investigate the impact of adding LoRA to different positions within the Transformer block. A typical Transformer block consists of the query, key, and value (QKV) linear layers, the output linear layer (O) in the multi-head attention module, and the two linear layers in the feedforward network (FFN). Our analysis, presented in Table \ref{tbl:ablation_lora}, demonstrates that applying LoRA to all of these linear layers results in the best overall performance. 

% \begin{table}[h]
% \centering
% \scalebox{1}{
% \begin{tabular}{lrrr}
% \toprule
% LoRA Target Modules &   OP  & FWT & BWT \\
% \midrule
% \rule{0pt}{4pt} Attention Q V  & 45.1 & 1.3  & -2.2  \\

% \rule{0pt}{8pt} Attention Q K V O & 45.3  & 1.4 & -2.3 	   \\
% \rule{0pt}{8pt} FFN  & 49.8   &  1.0 & -1.9 	   \\
% \rule{0pt}{8pt} Attention All + FFN & 45.7  &2.1 & 	-2.0    \\

% \bottomrule
% \end{tabular}}
% \caption{Ablation study on LoRA target modules, using T5-large as the backbone.}
% \label{tbl:ablation_lora}
% \end{table}

\subsection{Time Complexity Analysis}
\label{sec:time}
In this section, we discuss the time complexity challenges introduced by multi-round merging.
Generally, multi-round merging methods tend to have higher time complexity than traditional merging approaches. To mitigate this, we optimized the time complexity during the design of our framework. Our forgetting signal requires monitoring the loss changes of historical data. To reduce complexity, we insert historical data into the new data batch during implementation, performing only loss calculation without updating gradients, thus avoiding additional training costs.

Furthermore, in our merging method, we directly use the parameter change between two successive merges, rather than merging the entire model before and after training, which further improves the efficiency of merging.
Quantitatively, we compare the training time of our method with that of LoRAReplay, the single-round merging method TaSL, and Recurrent-KIF, which also performs multiple merges. The results are shown in Tablee~\ref{tbl:time}.

The results indicate that although our method takes slightly more time than LoRA Replay and TaSL, with an average increase of approximately 1.3 times, it delivers significant performance improvements. Moreover, compared to Recurrent-KIF, which also uses multi-round merging, our method benefits from adaptive merging timing, filtering out many unnecessary merges, and achieves lower time complexity through model design optimizations.

\begin{table}[t]
\centering
\scalebox{0.6}{
\begin{tabular}{lrrrr}
\toprule
Training Time \\ (Min/Epoch) & T5-large & Qwen3-1.7B & LLaMA2-7B & LLaMA2-13B  \\
\midrule

\rule{0pt}{4pt}LoRAReplay	   &1.4 & 3.3 & 4.5 & 6.6	 \\
\rule{0pt}{8pt}TaSL	   &1.4 & 3.4 & 4.6 & 6.7	 \\
\rule{0pt}{8pt}Recurrent-KIF   &1.4 & 4.9  & 5.5 & 9.1 \\
\rule{0pt}{8pt}{\ouralg}   &1.4 & 4.4 & 5.9 & 8.5 \\
\bottomrule
\end{tabular}}
\caption{Training time comparison across backbones.}
\label{tbl:time}
\end{table}

% \subsection{Sensitivity Analysis for Hyperparameters}
% The proposed framework incorporates three key hyperparameters: the initial iteration interval, the sliding window length, and the scaling factor for the forgetting signal threshold. Our analysis aims to evaluate the impact of varying these hyperparameters on the performance of our method, using the T5-large backbone model for testing.

\subsection{Generalizability of Learning and Forgetting Signals}
To validate the robustness of our method in highly imbalanced or severely shifting environments, we further conducted cross-dataset experiments, combining 4 tasks from the Standard CL benchmark and 15 from the SuperNI Benchmark, and tested on a 19-task sequence. The results are shown in the table \ref{tab:cross_dataset}.

\begin{table}[t]
\centering
\scalebox{0.9}{
\begin{tabular}{lccc}
\toprule
Method & OP $\uparrow$ & FWT $\uparrow$ & BWT $\uparrow$ \\
\midrule
Replay              & 37.7  & -13.5 & -21.8 \\
VR-MCL              & 44.9  &  -6.1 & -15.7 \\
Recurrent-KIF       & 46.4  &  -5.9 & -14.6 \\
AimMerging (ours)   & \textbf{48.3}  & \textbf{-3.0} & \textbf{-9.1} \\
Multi-task Learning & 57.2 & - & - \\
\bottomrule
\end{tabular}}
\caption{Cross-dataset evaluation on a 19-task sequence (4 Standard CL tasks + 15 SuperNI tasks). AimMerging consistently outperforms baselines.}
\label{tab:cross_dataset}
\end{table}

As shown in the table, our method still outperforms other baseline methods, with an average improvement of 1.9\% on OP, 2.9\% on FWT, and 5.5\% on BWT compared to Recurrent-KIF. These results will be included in the revised paper.

\subsection{Sensitivity Analysis of Hyperparameters}
Our method involves several hyperparameters in the learning and forgetting signals. Specifically, in the learning signal we consider the initial merging interval $S_{init}$, the sliding window size $L_w$, and the range for merging intervals $S_{min}, S_{max}$; while in the forgetting signal, we consider the threshold scaling factor $\gamma_{forget}$ and the maximum activation count $F_{max}$. To evaluate the sensitivity of our method to these hyperparameters, we conducted experiments on the Standard CL benchmark (task order 1) using the T5-large backbone model. The results are presented in Table~\ref{tab:hyper_sensitivity}.

\begin{table}[t]
\centering
\scalebox{0.85}{
\begin{tabular}{lcccc}
\toprule
Hyperparameter & Value & OP $\uparrow$ & FWT $\uparrow$ & BWT $\uparrow$ \\
\midrule
\multirow{4}{*}{$S_{init}$} 
  & 2   & 78.2 & -1.5 & -0.7 \\
  & 8   & 78.3 & -1.4 & -0.6 \\
  & 16  & 78.0 & -1.7 & -1.0 \\
  & 32  & 77.7 & -2.0 & -1.5 \\
\midrule
\multirow{4}{*}{$L_w$} 
  & 2   & 78.3 & -1.5 & -0.6 \\
  & 3   & 78.3 & -1.4 & -0.6 \\
  & 4   & 78.4 & -1.3 & -0.7 \\
  & 8   & 78.5 & -1.5 & -0.7 \\
\midrule
\multirow{4}{*}{$S_{min}, S_{max}$} 
  & 2, 128  & 78.3 & -1.4 & -0.6 \\
  & 8, 128  & 78.0 & -1.6 & -0.9 \\
  & 8, 64   & 78.2 & -1.6 & -0.7 \\
  & 2, 64   & 78.5 & -1.5 & -0.7 \\
\midrule
\multirow{4}{*}{$\gamma_{forget}$} 
  & 2   & 78.3 & -1.4 & -0.6 \\
  & 8   & 78.1 & -1.5 & -0.8 \\
  & 16  & 78.0 & -1.7 & -0.9 \\
  & 32  & 77.6 & -1.9 & -1.4 \\
\midrule
\multirow{4}{*}{$F_{max}$} 
  & 2   & 78.5 & -1.3 & -0.8 \\
  & 3   & 78.3 & -1.4 & -0.6 \\
  & 4   & 78.4 & -1.5 & -0.7 \\
  & 8   & 78.0 & -1.7 & -1.0 \\
\bottomrule
\end{tabular}}
\caption{Sensitivity analysis of key hyperparameters on the Standard CL benchmark (task order 1) with T5-large. Results show that the method remains robust when hyperparameters are set within reasonable ranges.}
\label{tab:hyper_sensitivity}
\end{table}

As shown in the table, increasing $S_{init}$ may lead to missing the optimal merging timing, resulting in more forgetting. Enlarging the sliding window size $L_w$ improves the stability of the learning signal, but overly long windows can accumulate erroneous data, causing performance degradation. Overall, when hyperparameters are set within reasonable ranges, the model remains robust and is not highly sensitive. This demonstrates that the lack of automated hyperparameter tuning does not compromise the practicality or reproducibility of our method.

\subsection{Comparison with Rehearsal-free Baselines}
Our method relies on memory data to obtain the forgetting signal, and thus directly removing the memory buffer is not straightforward. To ensure fairness in comparison, we additionally equipped prior rehearsal-free baselines with the same memory buffer and re-evaluated them (i.e., fine-tuning for two epochs on memory data after standard training). The results on the SuperNI benchmark (task order 6) with the T5-large backbone are presented in Table~\ref{tab:rehearsal_free}.

\begin{table}[t]
\centering
\scalebox{0.9}{
\begin{tabular}{lccc}
\toprule
Method & OP $\uparrow$ & FWT $\uparrow$ & BWT $\uparrow$ \\
\midrule
Replay              & 35.6 & -12.7 & -15.4 \\
O-LoRA              & 39.2 &   0.2 &  -9.4 \\
MIGU                & 39.0 &  -1.1 &  -6.3 \\
TaSL                & 41.9 &  -0.4 &  -5.9 \\
MoCL                & 42.4 &  -0.3 &  -5.8 \\
AimMerging (ours)   & \textbf{44.1} & \textbf{2.3} & \textbf{-3.9} \\
\bottomrule
\end{tabular}}
\caption{Comparison with rehearsal-free baselines on the SuperNI benchmark (task order 6) using T5-large. All methods benefit from memory buffer usage, but AimMerging achieves the best performance across all metrics.}
\label{tab:rehearsal_free}
\end{table}

As shown in the results, all baselines benefit from the memory buffer. However, AimMerging still consistently achieves the best performance across OP, FWT, and BWT. This demonstrates that our method retains its advantage even under comparable settings. We leave the development of a memory-free variant of AimMerging as an important direction for future work.

\section{Dataset Statistics}
\label{sec:dataset}
We adopt the experimental setup from \citet{du2024unlocking}, using three CL benchmark datasets:
(i) \textbf{Standard CL Benchmark}, which consists of five text classification tasks from \citet{zhang2015character}: AG News, Amazon Reviews, Yelp Reviews, DBpedia, and Yahoo Answers.
(ii) \textbf{Long Sequence Benchmark}, a more challenging evaluation scenario comprising 15 tasks \cite{razdaibiedina2023progressive}: five from the Standard CL Benchmark, four from the GLUE benchmark \cite{wang2018glue}, five from SuperGLUE \cite{wang2019superglue}, and the IMDB Movie Reviews dataset \cite{maas2011learning}.
(iii) \textbf{SuperNI Benchmark}~\cite{wang2022super}, a comprehensive benchmark designed to evaluate a wide range of NLP tasks, includes tasks in dialogue generation \cite{xu2024parenting}, information extraction, question answering \cite{lu2021getting}, summarization \cite{hu2024longreciperecipeefficientlong}, and sentiment analysis \cite{xu2025dearllm, chen2025dontneedprebuiltgraphs}.

Table \ref{superni} \& \ref{long-sequence} show details of the datasets we used for our experiments, along with their evaluation metrics. Overall, in SuperNI \cite{chen2025prototype}, we choose 3 tasks from dialogue generation (Dialog) \cite{feng2024continual, dong2024zero}, information extraction (IE),  question answering (QA) \cite{zhao2024large}, summarization (Sum) and sentiment analysis (SA), respectively.

For the Long Sequence benchmark \cite{wang2024comprehensive}, this includes five tasks from the standard CL benchmark (AG News, Amazon reviews, Yelp reviews, DBpedia and Yahoo Answers), four from GLUE benchmark (MNLI, QQP, RTE, SST2), five from SuperGLUE benchmark (WiC, CB, COPA, MultiRC, BoolQ), and the IMDB movie reviews dataset \cite{feng2023towards2, chen2024entity}.

We report 7 different task orders used for our experiments in Table \ref{order}.

\begin{table*}
\centering
\scalebox{0.9}{
\begin{tabular}{lllll}
\toprule
\textbf{Dataset name} & \textbf{Task}  & \textbf{Metric} \\
\midrule
1. task639\_multi\_woz\_user\_utterance\_generation  & dialogue generation   & Rouge-L        \\
2. task1590\_diplomacy\_text\_generation & dialogue generation   & Rouge-L       \\
3. task1729\_personachat\_generate\_next & dialogue generation   & Rouge-L      \\
4. task181\_outcome\_extraction & information extraction & Rouge-L        \\
5. task748\_glucose\_reverse\_cause\_event\_detection & information extraction & Rouge-L       \\
6. task1510\_evalution\_relation\_extraction   & information extraction & Rouge-L  \\
7. task002\_quoref\_answer\_generation & question answering & Rouge-L \\
8. task073\_commonsenseqa\_answer\_generation & question answering & Rouge-L    \\
9. task591\_sciq\_answer\_generation  & question answering & Rouge-L        \\
10. task511\_reddit\_tifu\_long\_text\_summarization     & summarization        & Rouge-L        \\
11. task1290\_xsum\_summarization  & summarization       & Rouge-L        \\
12. task1572\_samsum\_summary  &summarization  & Rouge-L \\
13. task363\_sst2\_polarity\_classification  & sentiment analysis   & accuracy        \\
14. task875\_emotion\_classification & sentiment analysis   & accuracy  \\
15. task1687\_sentiment140\_classification & sentiment analysis   & accuracy  \\
\bottomrule
\end{tabular}}
\caption{The details of 15 datasets in the SuperNI Benchmark \cite{wang2022super}.
}
\label{superni}
\end{table*}

\begin{table*}[htbp]
\centering
\scalebox{0.9}{
\begin{tabular}{lllll}
\toprule
\textbf{Dataset name} & \textbf{Category} & \textbf{Task}             & \textbf{Domain}     & \textbf{Metric} \\ \midrule
1. Yelp               & CL Benchmark      & sentiment analysis        & Yelp reviews        & accuracy        \\
2. Amazon             & CL Benchmark      & sentiment analysis        & Amazon reviews      & accuracy        \\
3. DBpedia            & CL Benchmark      & topic classification      & Wikipedia           & accuracy        \\
4. Yahoo              & CL Benchmark      & topic classification      & Yahoo Q\&A          & accuracy        \\
5. AG News            & CL Benchmark      & topic classification      & news                & accuracy        \\
6. MNLI               & GLUE              & natural language
inference                       & various             & accuracy        \\
7. QQP                & GLUE              & paragraph detection       & Quora               & accuracy        \\
8. RTE                & GLUE              & natural language inference                       & news, Wikipedia     & accuracy        \\
9. SST-2              & GLUE              & sentiment analysis        & movie reviews       & accuracy        \\
10. WiC               & SuperGLUE         & word sense disambiguation & lexical databases   & accuracy        \\
11. CB                & SuperGLUE         & natural language
inference                       & various             & accuracy        \\
12. COPA              & SuperGLUE         & question and answering                        & blogs, encyclopedia & accuracy        \\
13. BoolQA            & SuperGLUE         & boolean question and answering                & Wikipedia           & accuracy        \\
14. MultiRC           & SuperGLUE         & question and answering                        & various             & accuracy        \\
15. IMDB              & SuperGLUE         & sentiment analysis        & movie reviews       & accuracy        \\ \bottomrule
\end{tabular}}
\caption{The details of 15 classification datasets in the Long Sequence Benchmark \cite{razdaibiedina2022progressive}. First five tasks
correspond to the standard CL benchmark \cite{zhang2015character}.
}
\label{long-sequence}
\end{table*}

\begin{table*}[h]
\centering
\scalebox{0.9}{
\begin{tabular}{lll}
\hline
\textbf{Order} & \textbf{Benchmark} & \textbf{Task Sequence}                                                                                                                                \\ \hline
1              &  \multirow{3}*{\tabincell{c}{Standard CL}}      & dbpedia → amazon → yahoo → ag                                                                                                                         \\
2              &       & dbpedia → amazon → ag → yahoo                                                                                                                         \\
3              &       & yahoo → amazon → ag → dbpedia                                                                                                                         \\ \hline
4              & \multirow{2}*{\tabincell{c}{\\ Long Sequence}}              & \begin{tabular}[c]{@{}l@{}}mnli → cb → wic → copa → qqp → boolqa → rte → imdb →\\ yelp → amazon → sst-2 → dbpedia → ag → multirc → yahoo\end{tabular} \\
%5              & T5             & \begin{tabular}[c]{@{}l@{}}multirc → boolqa → wic → mnli → cb → copa → qqp → rte\\ → imdb → sst-2 → dbpedia → ag → yelp → amazon → yahoo\end{tabular} \\
5              &              & \begin{tabular}[c]{@{}l@{}}yelp → amazon → mnli → cb → copa → qqp → rte → imdb →\\ sst-2 → dbpedia → ag → yahoo → multirc → boolqa → wic\end{tabular} \\ \hline
6              & \multirow{3}*{\tabincell{c}{\\ SuperNI}}  & \begin{tabular}[c]{@{}l@{}}task1572 → task363 → task1290 → task181 → task002 →\\ task1510 → task639 → task1729 → task073 → task1590 →\\ task748 → task511 → task591 → task1687 → task875\end{tabular} \\
7              &      & \begin{tabular}[c]{@{}l@{}}task748 → task073 → task1590 → task639 → task1572 →\\ task1687 → task591 → task363 → task1510 → task1729 →\\ task181 → task511 → task002 → task1290 → task875\end{tabular} \\ \hline
\end{tabular}}
\caption{Seven different orders of task sequences used for continual learning experiments. Orders 1-3 correspond to the standard CL becnhmark adopted by prior works. Orders 4-5 are long-sequence orders spanning 15 tasks, and orders 6-7 are superni spanning 15 tasks following \cite{razdaibiedina2023progressive}.}
\label{order}
\end{table*}

\section{Implementation Details}
\label{sec:details}
Experiments are implemented using PyTorch and the Transformer library, running on 8 NVIDIA V100 GPUs with 32GB memory. The following hyperparameters are used:
a learning rate of 3e-4, a batch sizes of 8, and training for 10 epochs. The LoRA settings are: $r = 8$, $\alpha = 32$, dropout = 0.05, targeting modules [q\_proj,v\_proj]. For testing: temperature = 0.02, top\_p = 0, top\_k = 1, num\_beams = 1, max new tokens = 128.

It is worth noting that we used the same hyper-parameters across different datasets and backbones, demonstrating the generalizability of our method without requiring extensive hyperparameter tuning for each specific setting.

\end{document}